\pdfoutput=1

\documentclass[11pt]{article}

\usepackage[]{EMNLP2023}

\usepackage{times}
\usepackage{latexsym}

\usepackage[T1]{fontenc}

\usepackage[utf8]{inputenc}

\usepackage{microtype}

\usepackage{inconsolata}

\usepackage{verbatim}

\definecolor{cornellred}{rgb}{0.7, 0.11, 0.11}

\definecolor{dartmouthgreen}{rgb}{0.05, 0.5, 0.06}

\definecolor{vibrantorange}{rgb}{1.0, 0.55, 0.0}
\definecolor{darkerorange}{rgb}{0.8, 0.4, 0.0}

\definecolor{turquoise}{RGB}{64,224,208}

\definecolor{darkturquoise}{RGB}{64,224,208}

\definecolor{highlightcolor}{RGB}{102, 255, 178} 

\newlength{\originalfboxsep}
\newcommand{\highlight}[1]{%
  \setlength{\fboxsep}{1pt}
  \colorbox{highlightcolor}{#1}%
  \setlength{\fboxsep}{\originalfboxsep}
}

\definecolor{maxdevpuple}{HTML}{64a216}

\definecolor{ptruecolor}{HTML}{e9c46a}
\definecolor{ourscolor}{HTML}{1a80bb}

\definecolor{epochsred}{HTML}{9b2226}

\definecolor{correctcolor}{RGB}{0, 109, 119}
\definecolor{wrongcolor}{RGB}{226, 150, 120}

\definecolor{categorycolor}{RGB}{31, 78, 121}
\definecolor{modelcolor}{RGB}{0, 0, 0}

\newcommand{\score}{P_{\mathtt{Correct}}}

\newcommand{\method}{{\rmfamily \textbf{S}li\textbf{CK}}\xspace}

\newcommand{\knownSolo}{$\mathtt{Known}$\xspace}

\newcommand{\knowncat}{$\mathtt{HighlyKnown}$\xspace}

\newcommand{\maybeknowncat}{$\mathtt{MaybeKnown}$\xspace}
\newcommand{\weaklyknowncat}{$\mathtt{WeaklyKnown}$\xspace}
\newcommand{\unknowncat}{$\mathtt{Unknown}$\xspace}

\newcommand{\knowncatshort}{$\mathtt{Hkn}$}
\newcommand{\maybeknowncatshort}{$\mathtt{Mkn}$}
\newcommand{\weaklyknowncatshort}{$\mathtt{Wkn}$}
\newcommand{\unknowncatshort}{$\mathtt{Unk}$}

\newcommand{\known}{\textcolor{modelcolor}{$D_{\mathtt{HighlyKnown}}$}\xspace}
\newcommand{\maybeknown}{\textcolor{modelcolor}{$D_{\mathtt{MaybeKnown}}$}\xspace}
\newcommand{\weaklyknown}{\textcolor{modelcolor}{$D_{\mathtt{WeaklyKnown}}$}\xspace}

\newcommand{\random}{\textcolor{modelcolor}{$D_{\mathtt{Natural}}$}\xspace}

\newcommand{\unknownfull}{\textcolor{modelcolor}{$D_{\mathtt{Unknown}}$}\xspace}

\newcommand{\dknown}{\textcolor{modelcolor}{$D_{\mathtt{Known}}$}\xspace}

\newcommand{\didk}{\textcolor{modelcolor}{$D_{\mathtt{IDK}}$}\xspace}

\newcommand{\eq}{\textsc{\small{EntityQuestions}}\xspace}

\newcommand{\maxdev}{\textcolor{maxdevpuple}{\textsc{\small{early\_stop}}}\xspace}
\newcommand{\conv}{\textcolor{epochsred}{\textsc{\small{Convergence}}}\xspace}

\newcommand{\nl}[1]{``\textit{#1}''}

\newcommand{\M}{$M$\xspace}
\newcommand{\D}{$D$\xspace}
\newcommand{\MD}{$M_D$\xspace}

\usepackage{graphicx}
\usepackage{booktabs}
\usepackage{multirow}
\usepackage{xcolor}
\usepackage{colortbl}

\usepackage{xspace}
\usepackage[normalem]{ulem} 
\usepackage{amsmath} 
\usepackage{cleveref}
\usepackage{enumitem}
\usepackage{threeparttable} 
\usepackage{amssymb} 
\usepackage{boldline}
\usepackage{tabularx}
\usepackage{float}
\usepackage{soul}

\usepackage{adjustbox}

\usepackage{subcaption} 

\usepackage{bm}
\usepackage{mdframed}

\definecolor{darkgreen}{rgb}{0.0, 0.5, 0.0}

\title{
Does Fine-Tuning LLMs on New Knowledge Encourage Hallucinations?

}

\author{
Zorik Gekhman$^{T}$\Thanks{ Work done during an internship at Google Research.} \quad
Gal Yona$^{G}$ \quad 
Roee Aharoni$^{G}$\quad
Matan Eyal$^{G}$ \quad 
\textbf{Amir Feder}$^{G}$\quad \vspace{2mm}\\ 
\textbf{Roi Reichart}$^{T}$\quad 
\textbf{Jonathan Herzig}$^{G}$\quad \vspace{3mm}\\
$^{T}$Technion - Israel Institute of Technology \quad
$^{G}$Google Research \vspace{1mm}\\
{\tt zorikgekhman@gmail.com, jherzig@google.com} 
}

\begin{document}

    \maketitle
    \begin{abstract}

When large language models are aligned via supervised fine-tuning, they may encounter new factual information that was not acquired through pre-training.
It is often conjectured that this can teach the model the behavior of \emph{hallucinating} factually incorrect responses, as the model is trained to generate facts that are not grounded in its pre-existing knowledge. 
\mbox{In this work,} we study the impact of such exposure to new knowledge on the capability of the fine-tuned model to utilize its pre-existing knowledge.
To this end, we design a controlled setup, focused on closed-book QA, where we vary the proportion of the fine-tuning examples that introduce new knowledge.
We demonstrate that large language models struggle to acquire new factual knowledge through fine-tuning, as fine-tuning examples that introduce new knowledge are learned significantly slower than those consistent with the model's knowledge. 
However, we also find that as the examples with new knowledge are eventually learned, they linearly increase the model's tendency to hallucinate.
Taken together, our results highlight the risk in introducing new factual knowledge through fine-tuning, and support the view that 
large language models mostly acquire factual knowledge through pre-training, whereas fine-tuning teaches them to use it more efficiently.

\end{abstract}

    \section{Introduction}
\label{sec:intro}

\definecolor{myPurple}{RGB}{128, 0, 128}

\begin{figure}[t]
 \centering
  \vspace{-0.6cm}
 \includegraphics[width=\columnwidth]{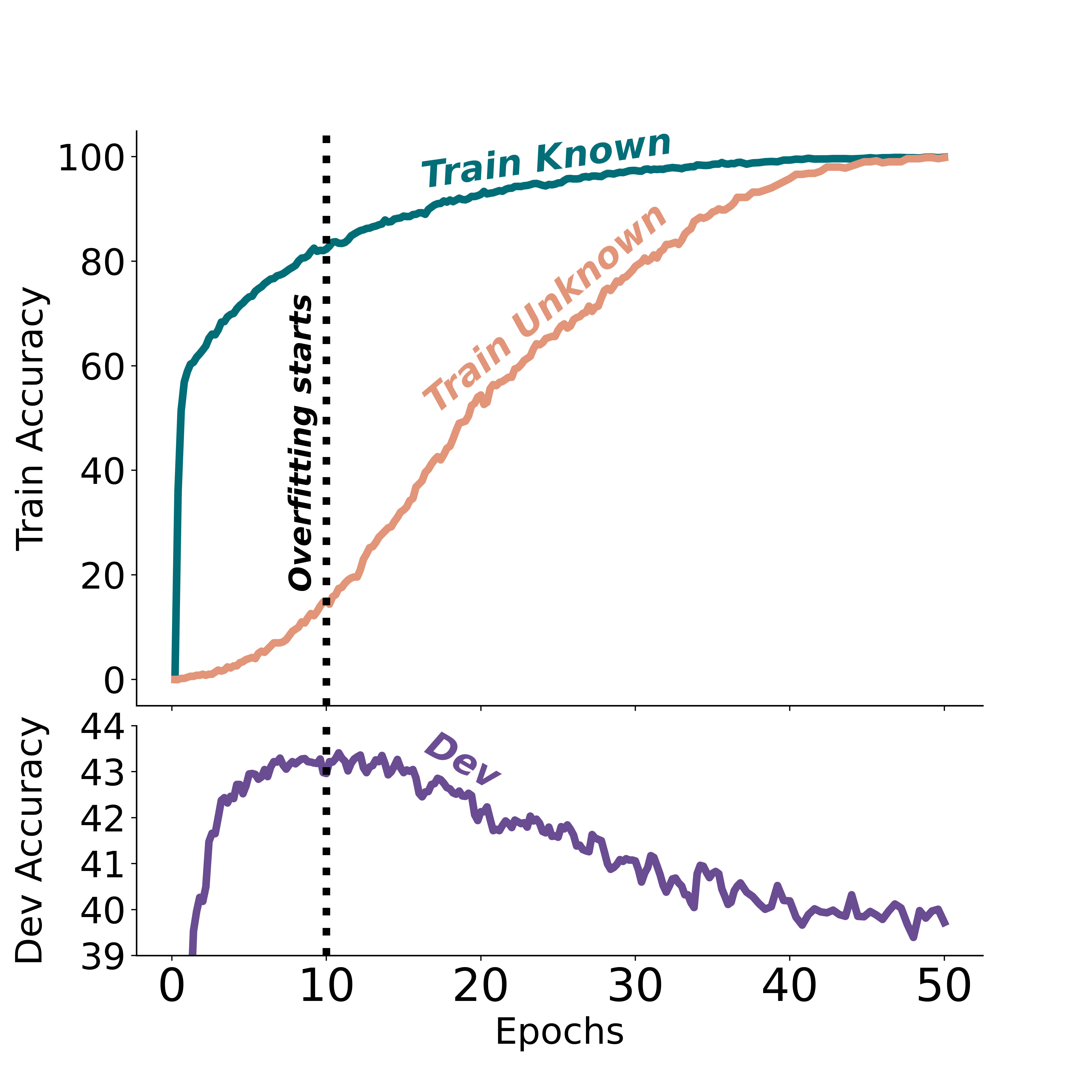}   
    \caption{
    Train and development accuracies as a function of the fine-tuning duration, when fine-tuning on $50\%$ \knownSolo and $50\%$ \unknowncat examples.
    \unknowncat examples are fitted substantially slower than \knownSolo. The best development performance is obtained when the LLM fits the majority of the \knownSolo training examples but only few of the \unknowncat ones. From this point, fitting \unknowncat examples reduces the performance.
    }
    \label{fig:main_plot}    
\end{figure}

Pre-training Large Language Models (LLMs) on textual corpora embeds substantial factual knowledge in their parameters \cite{LLMs_as_KG_1,LLMs_as_KG_4,LLMs_as_KG_3}, which is essential for excelling in various downstream applications.
These models often require further alignment to desired behaviors, typically achieved through supervised fine-tuning on instruction-following tasks \cite{Instruction-Tuning,NATURAL_INSTRUCTIONS} and preference learning from human feedback \cite{RL_OpenAI, rafailov2024direct}.

In the fine-tuning phase, the model is usually trained on outputs created by human annotators or other LLMs. As a result, the model may encounter new factual information, extending beyond the knowledge it acquired during pre-training.
This raises the question of how LLMs integrate new facts outside of their pre-existing knowledge. One possibility is that the model simply adapts by learning this new factual information. However,
a common conjecture posits that such exposure to new knowledge may encourage the model to \emph{hallucinate} factually incorrect responses, as the model is essentially trained to generate facts that are not grounded in its pre-existing knowledge \cite{Schulman_RL, hallucinations_survey, Behavior_Cloning_is_Miscalibrated, Goldberg_RL, The_False_Promise_of_Imitating_LLMs}.

In this work, we study how learning new factual knowledge through fine-tuning impacts the model's tendency to hallucinate w.r.t. its pre-existing knowledge, exploring the above conjecture.\footnote{
While we focus on supervised fine-tuning, our findings 
may be relevant to offline preference optimization methods like DPO~\cite{rafailov2024direct} that may add new knowledge, which we leave for future work.}

To study the impact of new knowledge, we must be able to assess whether a single fine-tuning example is consistent with the model's knowledge.
We propose \method, a hierarchy of four \emph{knowledge categories}, derived from a continuous 
measure that quantifies the agreement between model-generated answers and the ground-truth labels.
In \method, examples are first categorized into 
\knownSolo and \unknowncat types, 
where the latter corresponds to examples with facts that are most likely unknown to the model.
The \knownSolo examples are subsequently split into three categories: \knowncat, \maybeknowncat, and \weaklyknowncat
(\Cref{tab:categories_def}). 

Equipped with the above method, we carefully design a controlled study, focused on closed-book question answering (QA), where we vary the proportion of the fine-tuning examples categorized as \unknowncat, while controlling for other factors.

Our study empirically demonstrates that learning from \unknowncat fine-tuning examples is linearly correlated with the model's tendency to \emph{hallucinate} w.r.t. its pre-existing knowledge (\S \ref{sec:results}). Conversely, learning from \knownSolo examples is correlated with better utilization of pre-existing knowledge. 

Through an analysis of the training dynamics, we discover that the LLM fits \unknowncat fine-tuning examples \emph{substantially slower} than \knownSolo examples (top plot in \Cref{fig:main_plot}). 
This indicates that during fine-tuning, LLMs struggle\footnote{We use the term ``struggle'' to describe how LLMs converge slowly for examples containing new factual knowledge. Since this term carries emotional connotations, we note that we do not ascribe any emotional attributes to LLMs.} to integrate new factual knowledge (present in the \unknowncat fine-tuning examples). Instead, they mostly learn to expose their pre-existing knowledge (using the \knownSolo fine-tuning examples). 

From a practical perspective, mitigating overfitting using \emph{early-stopping} (vertical dotted line in \Cref{fig:main_plot}) can minimize the risk of the hallucinations caused by fitting the \unknowncat examples, since they primarily emerge in later training stages
as a form of overfitting (as illustrated by the development performance decline in the bottom plot of \Cref{fig:main_plot}).
Alternatively, we also show that \emph{filtering-out} the \unknowncat fine-tuning examples substantially reduces the risk of overfitting, without sacrificing performance.

We further evaluate the impact of fine-tuning examples from each of our three \knownSolo knowledge categories on performance (\S \ref{sec:single_cat}).
Unexpectedly, we find that a model fine-tuned only on examples from the highest knowledge degree, denoted \knowncat, does not yield the best results.
Our analysis reveals that incorporating \maybeknowncat fine-tuning examples, representing facts with lower degrees of certainty, plays an important part in properly handling such examples in test time.
This indicates that the composition of fine-tuning examples significantly influences the extent to which LLMs effectively utilize their pre-existing knowledge.

To summarize, we study the effect of new factual knowledge in the fine-tuning data by designing a controlled setup that isolates this factor. We find that fine-tuning examples that introduce new knowledge are learned slowly, which suggests that LLMs struggle to integrate new knowledge through fine-tuning and supports the view that LLMs mostly acquire knowledge through pre-training \cite{LIMA, The_Unlocking_Spell_on_Base_LLMs}. However, we also find that as the model eventually learns new knowledge through fine-tuning, it becomes more prone to hallucinations w.r.t. its pre-existing knowledge.
Collectively, our findings highlight the potential for unintended consequences when introducing new knowledge through fine-tuning, and imply that fine-tuning may be more useful as a mechanism to enhance the utilization of pre-existing knowledge.

    \section{Study Setup}
\label{sec:exp_setting}

\begin{figure*}[t]
    \centering
    
    \begin{subtable}{\textwidth}
        \centering
        \resizebox{\textwidth}{!}{%
        \begin{tabular}{|l|l|p{5.5cm}|p{11.3cm}|}
            \hline
              \multicolumn{1}{|c|}{Type}  &
              \multicolumn{1}{c|}{Category} &
              \multicolumn{1}{c|}{Definition} &
              \multicolumn{1}{c|}{Explanation} \\
              \hline
            
             \multirow{4}{*}{\knownSolo}& \multirow{1}{*}{\knowncat}   & $\score(q, a; M, T=0) = 1$ & Greedy decoding \emph{always} predicts the correct answer. \\

            \cline{2-4}
            
            & \maybeknowncat   &   $\score(q, a; M, T=0) \in (0,1) $ & Greedy decoding \emph{sometimes} (but not always) predicts the correct answer. \\
            
            \cline{2-4}
            
             & \multirow{2}{*}{\weaklyknowncat}  &  $\score(q, a; M, T=0)=0 \land \score(q, a; M, T>0)>0$ & Greedy decoding \emph{never} predicts the correct answer, whereas temperature sampling with $T > 0$ \emph{sometimes} predicts the correct answer. \\
             
            \hline

            \multirow{2}{*}{\unknowncat} & \multirow{2}{*}{\unknowncat} &  \multirow{2}{*}{$\score(q, a; M, T\geq0)=0$} & 
            The model \emph{never} predicts the correct answer, thus it seem to lack the knowledge of the correct answer. \\
            \hline

            \end{tabular}%
        }
        \caption{
        }
        \label{tab:categories_def_definitions}
    \end{subtable}
    
    \vspace{0.5em} 
    
    \begin{subtable}{\textwidth}
        \centering
        \resizebox{\textwidth}{!}{%
        \begin{tabular}{|l|l|l|l|l|}
            \hline
            
              \multicolumn{1}{|c|}{Category} &
              \multicolumn{1}{c|}{Question} &
              \multicolumn{1}{c|}{Gold Answer} &
              \multicolumn{1}{c|}{Greedy Answers} &
              \multicolumn{1}{c|}{Sampled Answers} \\
              \hline
            
            \knowncat   & \textit{Who founded Science of Mind?} & \textit{\highlight{Ernest Holmes}}
             & [\textit{\highlight{Ernest Holmes}, .. \highlight{Ernest Holmes}, ..}] & [\textit{..., ...}] \\
            
            
            \maybeknowncat   &   \textit{What is the capital of Toledo District?} & \textit{\highlight{Punta Gorda}} & [\textit{Belmopan, .., \highlight{Punta Gorda}, ..}] & [\textit{..., ...}] \\
            
            \weaklyknowncat  &  \textit{What kind of work does Scott McGrew do?} & \textit{\highlight{Journalist}}
             & [\textit{Film director, .. Actor, ..}] & [\textit{Musician, .. \highlight{Journalist}, ..}] \\

            \unknowncat &  \textit{Where is Benedict located?
            } & \textit{\highlight{Hubbard County}} & [\textit{Louisiana, .. New Mexico, ..}] & [\textit{Washington, .. Texas, ..}]  \\

            \hline
            \end{tabular}%
        }
        \caption{
        }
        \vspace{-0.5em} 
        \label{tab:categories_def_examples}
    \end{subtable}
    
    \caption{
    Formal definitions of the \method knowledge categories, based on the $\score$ measure as defined in \S \ref{sec:categorizing} \textbf{(a)}, accompanied with real examples from the annotated \eq dataset  used in our study \textbf{(b)}.
}
\label{tab:categories_def}
\end{figure*}

Given a fine-tuning dataset \D and a pre-trained LLM \M, we denote by \MD a model obtained by fine-tuning $M$ on $D$.
To study how new knowledge in $D$ affects $M_D$'s performance, we design a controlled setup creating variants of $D$ with varying proportions of examples that are unknown to $M$.

When constructing $D$, our objective is to reflect instruction tuning on diverse knowledge-intensive tasks while maintaining control over the experimental setting.
We thus focus on factual knowledge that can be structured as \emph{(subject, relation, object)} triplets, which are converted into closed-book QA format. In this setup, $D=\{(q_i,a_i)\}_{i=1}^N$, where $q$ is a knowledge-seeking question  corresponding to a specific triplet (e.g., \nl{Where is Paris located?}) and $a$ is the ground-truth answer (e.g., \nl{France}). 
To this end, we use \eq \cite{Entity_Questions}, where triplets from a diverse set of relations from Wikidata \cite{Wikidata} are converted to QA pairs. These relations encompass a broad spectrum of factual knowledge, including biographical information, geographical data, ownership and authorship details, history and more.
We use the original development and test splits, and we sub-sample the train split to create different variants of $D$. We focus on 12 diverse relations and reserve 7 additional relations for an \emph{out-of-distribution} test set, used (only) in \S \ref{sec:ood}.

As $M$, we use the PaLM 2-S base model\footnote{PaLM-2 is available in five sizes: XXS, XS, S, M, L, with the S version representing the middle size in this range.} \cite{PaLM2}. We focus on exact match (EM) as our evaluation metric.\footnote{We validated that in our setting EM strongly correlates with word-level F1 \cite{SQUAD}, and we choose EM as it is more intuitive for the purposes of our analysis.} Full technical details are in \S \ref{sec:data_prep_appendix}.

    \section{Quantifying Knowledge in LLMs}
\label{sec:categorizing}

To assess the effect of new knowledge in $D$ on the performance of $M_D$, we have to annotate each $(q,a)$ pair in $D$ w.r.t. whether $M$ knows that the answer to $q$ is $a$.\footnote{We also considered using \emph{fake} facts for introducing new knowledge, but we were concerned that this would introduce confounding factors into our study, as fake facts may behave differently than real ones. We discuss this in detail in \S \ref{sec:fake_facts_appendix}.}
To estimate this, we define a continuous $\score$ measure based on samples from $M$, and use it to divide $(q,a)$ pairs into four \emph{knowledge categories}. We name this approach \method (\textbf{S}ampling-based \textbf{C}ategorization of \textbf{K}nowledge).

\begin{figure*}[t]
 \centering
 \begin{subfigure}{.5\textwidth}
   \centering
   \vspace{-1.7mm} 
\includegraphics[width=\linewidth]{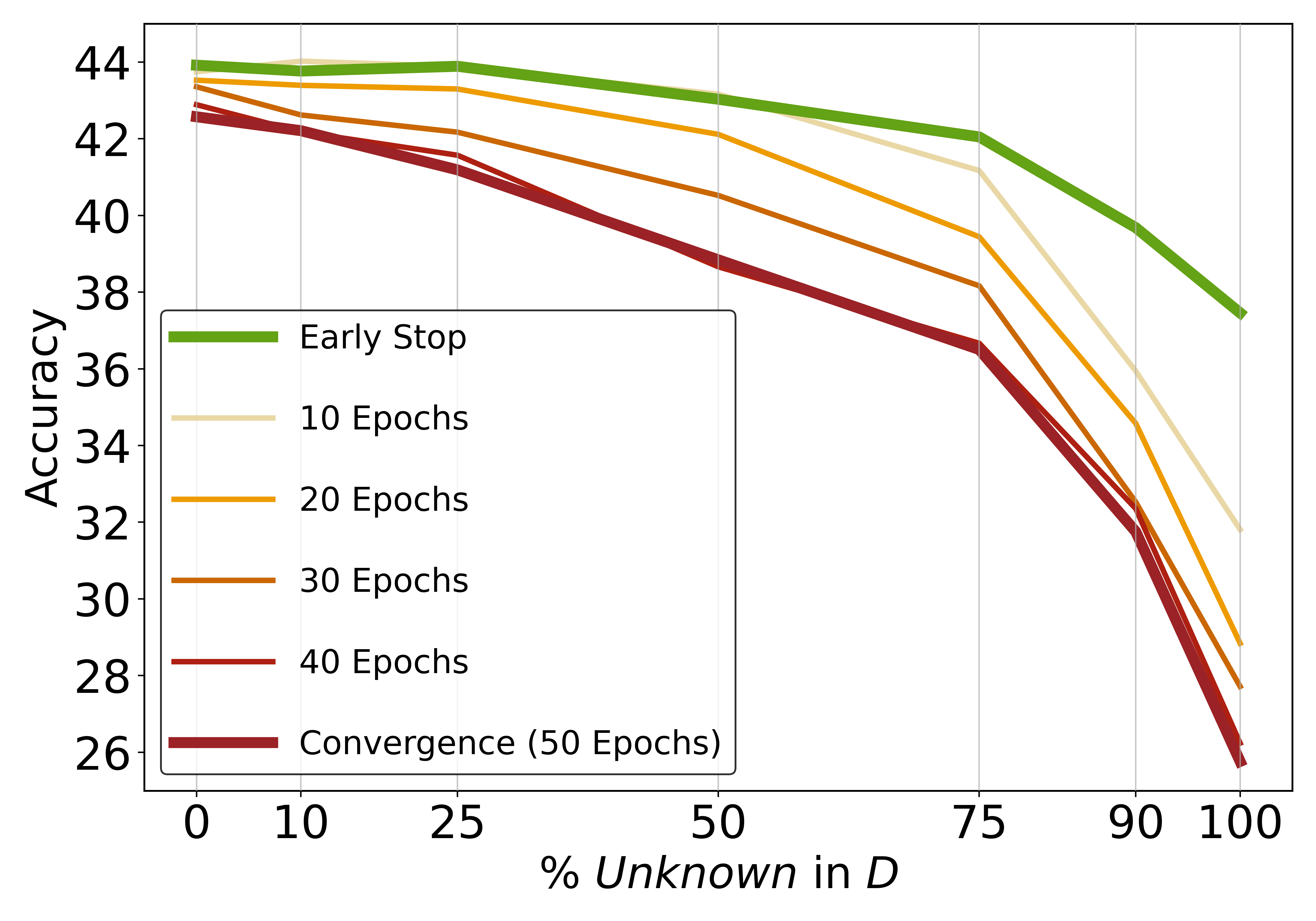}
    \vspace{-1.8em} 
   \caption{
   }
   \label{fig:fixed_steps_plot}
 \end{subfigure}%
 \begin{subfigure}{.5\textwidth}
   \centering
   \vspace{-1.7mm} 
\includegraphics[width=\linewidth]{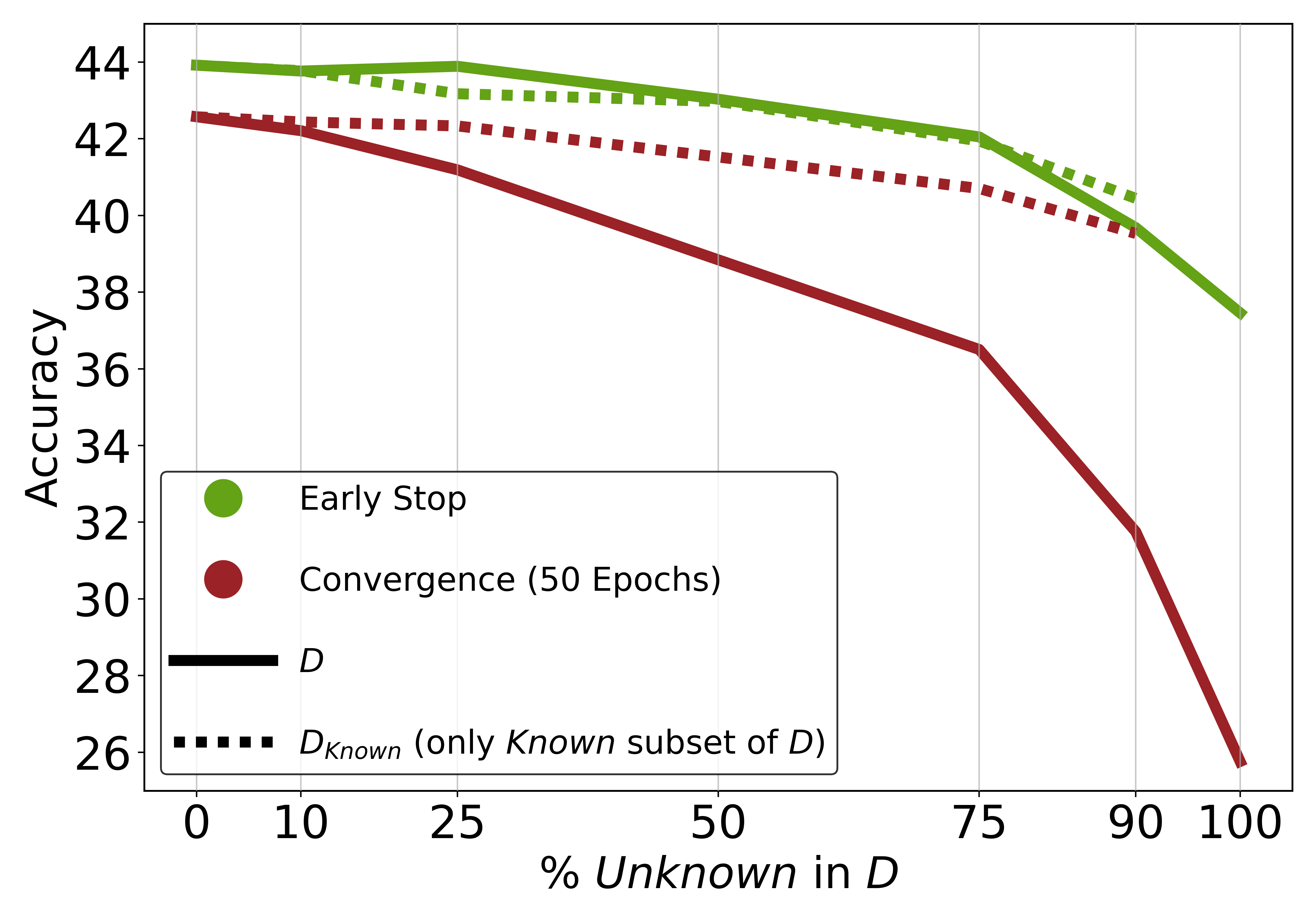}
  \vspace{-1.8em} 
   \caption{
   }
   \label{fig:ablated_plot}
 \end{subfigure}
 \caption{
  Test performance as a function of the $\%$ of \unknowncat examples in the fine-tuning dataset $D$.
 In \textbf{(a)}, each line corresponds to a different (fixed) number of epochs, except the \maxdev, which corresponds to early-stopping using the development set (see \S \ref{sec:results}). In \textbf{(b)} we present the ablation from \S \ref{sec:harmful_or_neutra}. Full lines correspond to fine-tuning on $D$ and are identical to (a). Dotted lines correspond to fine-tuning on the ablated variants \dknown, where \unknowncat examples are filtered-out. For $0\%$ \unknowncat $D=$\dknown and for $100\%$ \unknowncat there is no \dknown.
 }
 \label{fig:combined}
\end{figure*}

\paragraph{Defining $\boldsymbol{P_{\boldsymbol{\mathtt{Correct}}}}$.} 
We adopt the perspective that $M$ \emph{knows} that the answer to $q$ is $a$ if it generates $a$ when prompted to answer $q$ \cite{p_true,manakul2023selfcheckgpt}. Since $M$ is a base model that has not been specifically fine-tuned to follow instructions, we prompt $M$ using in-context learning with few-shot exemplars. Following \citet{Learning_To_Retrieve_Prompts}, we make sure that the few-shot exemplars have high semantic similarity to $q$.\footnote{In our study we achieve this by using exemplars from the same relation. E.g., if $q=$\nl{Where is Paris located?}, the exemplars would follow the pattern \nl{Where is \{X\} located?}.} 

In practice, $M$ can predict different answers since (1) the choice of exemplars influences individual predictions and (2) temperature sampling, if used, introduces randomness.
To reflect this, we define $\score(q, a; M, T)$ as an estimate of how likely is $M$ to accurately generate the correct answer $a$ to $q$, when prompted with \emph{random few-shot exemplars} and using decoding temperature $T$.

For the purposes of our study we approximate the value of $\score$ using $N_{\text{ex}}=10$ different random 4-shot prompts.\footnote{We use 4-shot simply since we found it enough for $M$ to output answers in the correct format.} For each \mbox{4-shot} prompt, we predict the greedy answer using $T=0$ and $16$ sampled answers using $T=0.5$.
$\score(q, a; M, T=0)$ is estimated by the fraction of correct greedy answers, and $\score(q, a; M, T>0)$ by the fraction of correct sampled answers. Full details are in \S \ref{sec:score_measure_appendix}.

\paragraph{Deriving knowledge categories from $\boldsymbol{P_{\boldsymbol{\mathtt{Correct}}}}$.} 
We define the \unknowncat category (bottom row in Figures \ref{tab:categories_def_definitions} and \ref{tab:categories_def_examples}) to represent $(q,a)$ pairs for which $M$ \emph{never} predicts the correct answer to $q$. In our notations this means that $\score(q, a; M, T\geq0)=0$. 
Alternatively, if $\score(q, a; M, T\geq0)>0$, i.e. $M$ \emph{sometimes} predicts the correct answer to $q$, we consider $(q,a)$ as \knownSolo. In this choice, we posit that if prompting $M$ to answer $q$ can \emph{sometimes} result with the correct answer $a$, then $M$ must have some association with the relevant fact.

Recognizing that knowledge can vary in degrees of certainty and extent, we divide the \knownSolo $(q,a)$ pairs into three distinct categories (top three rows in Tables \ref{tab:categories_def_definitions} and \ref{tab:categories_def_examples}). Motivated by the principle that $M$ should \emph{consistently} predict $a$ if $(q,a)$  is \knownSolo, we put emphasis on \emph{greedy decoding} outcomes, represented with $\score(q, a; M, T=0)$. \knowncat represents $(q,a)$ pairs for which $M$ \emph{always} greedily predicts $a$. If $M$ \emph{sometimes} (but not always) greedily predicts $a$, we consider $(q,a)$ as \maybeknowncat. Lastly, if $M$ \emph{never} greedily predicts $a$, we classify $(q,a)$ as \weaklyknowncat.

We apply \method to annotate each $(q,a)$ pair in our dataset with its knowledge category w.r.t. $M$.\footnote{In \eq we have $24\%$ \knowncat, $23\%$ \maybeknowncat, $17\%$, \weaklyknowncat, and $36\%$ \unknowncat. Full per-relation statistics are in \S \ref{sec:data_annotation_appendix}.} We analyze the quality of our categories in \S \ref{sec:taxonomy}.

    \section{How Harmful are \unknowncat Examples?}
\label{sec:results}

In this section we study the effect of new knowledge in the fine-tuning dataset $D$ on performance.
To isolate this effect, we vary the proportion of \unknowncat examples in $D$, while controlling for other factors. Specifically, we fix $|D|$ 
and create variants of $D$ with $X\%$ of \unknowncat and $(100-X)\%$ \knownSolo examples (full details in \S \ref{sec:controlled_experiment_appendix}). We
treat the \knownSolo categories collectively (see \Cref{tab:categories_def_definitions}), and provide a per-category analysis in \S \ref{sec:single_cat}. We denote early-stopping based on the development set as \maxdev (happens after 5-10 epochs) and 50 fine-tuning epochs as \conv, as at this point $M$ always completely fits $D$ (i.e. $100\%$ training accuracy). 
We measure test performance as a proxy for hallucinations since we are in a closed-book QA setup with disjoint train/test splits, where the model has to use its per-existing knowledge to answer test questions (see \S \ref{sec:hallucinations_vs_perfromance_appendix} for further discussion).

\subsection{Higher \unknowncat Ratio is Proportional to Performance Degradation}
\label{sec:higher_unk_affects_performance}

\Cref{fig:fixed_steps_plot} presents the performance as a function of the \% of \unknowncat examples in $D$, for different fine-tuning durations. Higher \%\unknowncat leads to performance degradation, regardless of the fine-tuning duration, which indicates that \unknowncat examples are less useful than \knownSolo.
Performance is also strongly affected by the fine-tuning duration, with \textcolor{maxdevpuple}{\maxdev} typically yielding the best performance. Training for more epochs usually reduces performance (with the lowest performance observed for \conv), which can be attributed to overfitting $D$. Interestingly, this effect increases with larger $\%$\unknowncat (the inter-line spacing from \maxdev exhibits a monotonic increase along the positive x-axis), suggesting that a higher \%\unknowncat increases the risk of overfitting.

\subsection{\unknowncat Examples: Harmful or Neutral?}
\label{sec:harmful_or_neutra}

Since $|D|$ is fixed, performance drops for higher \%\unknowncat could stem from simply the lower number of the \knownSolo fine-tuning examples. Thus, it is still not clear if \unknowncat examples are \emph{harmful} or \emph{neutral}.
To address this, we measure the effect of filtering-out all the \unknowncat examples from $D$. For each $D$ variant, we create a corresponding ablated variant \dknown, consisting only from the \knownSolo examples in $D$. E.g., if $D$ has $25\%$ \unknowncat, we filter them out and are left with the remaining $75\%$ \knownSolo examples and get $|$\dknown$|=0.75\times|D|$.

\Cref{fig:ablated_plot} presents the results. Perhaps surprisingly, for \textcolor{maxdevpuple}{\maxdev} the results for $D$ are almost identical to \dknown, indicating that the \unknowncat examples had a \emph{neutral} effect on performance (as their removal had minimal impact). Conversely, the \textcolor{epochsred}{\conv} results show that with longer training, \unknowncat examples are actually very \emph{harmful}. In this case $D$ under-performs \dknown, and the gap between them is proportional to the \unknowncat ratio. 

Interestingly, for \dknown, the gap between \textcolor{maxdevpuple}{\maxdev} and \textcolor{epochsred}{\conv} is very small (dotted lines), while this gap is very large for $D$ (full lines). 
This indicates that the presence of \unknowncat examples is what makes the variants with higher \unknowncat ratios more prone to overfitting.

\subsection{\unknowncat Examples are Fitted Slower than \knownSolo Examples}
\label{sec:ignore}

\begin{figure}[t]
 \centering
\vspace{-1.5mm} 
 \includegraphics[width=\columnwidth]{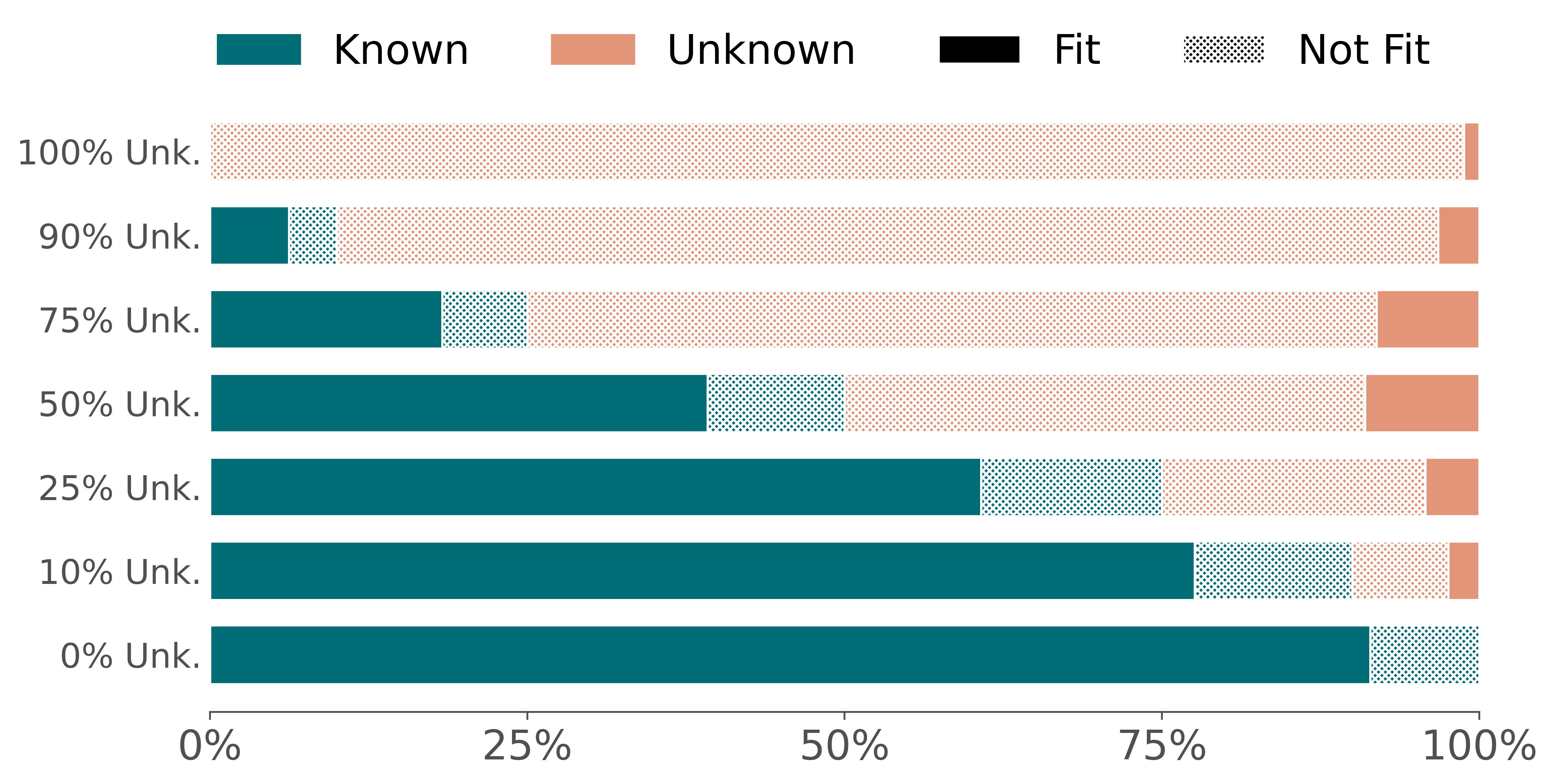}   
    \caption{
    The state of the examples in the fine-tuning dataset $D$ after \maxdev. For each variant of $D$ (y-axis), we illustrate which portion of the examples in $D$ the model fits (i.e. predicts the correct answer for $q$).
    }
    \label{fig:new_pies}    
\end{figure}

We showed that \unknowncat examples are harmful, but their negative effect is mostly materialized in later training stages, and thus can be empirically avoided using early stopping. To better understand these trends, we analyze the training dynamics by examining which fine-tuning examples in $D$ were fitted by $M$ during various fine-tuning stages. \Cref{fig:main_plot} presents the train accuracy of the \knownSolo and \unknowncat subsets of $D$ as a function of the fine-tuning duration. The development accuracy is presented in a zoomed-in plot at the bottom, as it falls within a narrower range. We include a breakdown of the train accuracy per \knownSolo category in \S \ref{sec:ltraining_accuracy_appendix}.

$M$ fits \unknowncat fine-tuning examples substantially slower than \knownSolo. 
In \maxdev (vertical dotted line), $M$ reaches peak performance on the development set, while fitting the majority of the \knownSolo examples but only a small fraction of the \unknowncat.
In \Cref{fig:new_pies}, we show that this behavior is consistent across all our variants of $D$. 
This can explain why in \maxdev the \unknowncat examples had a \emph{neutral} effect on performance (\S \ref{sec:harmful_or_neutra}), as at this point $M$ still did not fit most of them. 
Lastly, since \unknowncat examples are the ones that are likely to introduce new factual knowledge, their significantly slow fitting rate suggests
that LLMs struggle to acquire new factual knowledge through fine-tuning, instead they learn to expose their pre-existing knowledge using the \knownSolo examples.

\subsection{The Influence of \unknowncat vs \knownSolo
on Accuracy: A Linear Model Perspective}
\label{sec:equation}


\begin{table}[t]
\centering
\resizebox{\columnwidth}{!}{%
\begin{tabular}{lllll}

\multicolumn{1}{c}{} &
  \multicolumn{1}{c}{\text{$\beta_0$}} &
  \multicolumn{1}{c}{\text{$\beta_\text{kn}$}} &
  \multicolumn{1}{c}{\text{$\beta_\text{unk}$}} &
  \multicolumn{1}{c}{\text{$R^2$}} \\
\toprule

\multicolumn{1}{l}{In-distribution (\S \ref{sec:equation})}   & $36.9$ & $7.3$
 & $-8.3$ & $0.86$ \\

\multicolumn{1}{l}{Out-of-distribution (\S \ref{sec:ood})}     & $36.2$ & $3.2$
 & $-3.0$ & $0.95$ \\

\bottomrule
\end{tabular}%
}
\caption{
Results of our linear model for predicting the test accuracy as defined by \Cref{eq:accuracy}.
}
\label{tab:linear_model}
\end{table}

\begin{table*}[t]
\centering
\resizebox{\textwidth}{!}{%
\begin{tabular}{lrcrrrrcrcrrrr}

\multicolumn{1}{c}{\multirow{1}{*}{}} &
\multicolumn{6}{c}{\maxdev} &
\phantom{abcd} &
\multicolumn{6}{c}{\conv} \\

\cmidrule{2-7} \cmidrule{9-14} 

\multicolumn{1}{c}{\multirow{1}{*}{}} &
  \multicolumn{1}{c}{$\mathtt{Full}$} &
\phantom{a}&
  \multicolumn{1}{c}{\knowncatshort} &
  \multicolumn{1}{c}{\maybeknowncatshort} &
  \multicolumn{1}{c}{\weaklyknowncatshort} &
  \multicolumn{1}{c}{\unknowncatshort} &&
  \multicolumn{1}{c}{$\mathtt{Full}$} &
  \phantom{a}&
  \multicolumn{1}{c}{\knowncatshort} &
  \multicolumn{1}{c}{\maybeknowncatshort} &
  \multicolumn{1}{c}{\weaklyknowncatshort} &
  \multicolumn{1}{c}{\unknowncatshort} \\
\toprule


\known & 40.5  && \textbf{98.7} & 60.1 & 9.0 & 0.6 && 40.0 &&  \textbf{98.4} & 58.8 & 8.5 & 0.7 \\

\maybeknown & \textbf{43.6}  && \textbf{98.4} & \textbf{69.9} & 12.1 & 1.0 && \textbf{43.2}  && 97.5 & \textbf{68.2} & 12.9 & 1.3 \\

\weaklyknown & 39.2  && 95.0 & 59.2 & 8.6 & 0.4 && 35.4  && 73.5 & 55.8 & \textbf{17.2} & 2.2 \\

\unknownfull & 37.5 &&  95.6 & 52.9 & 6.5 & 0.6 && 25.8  && 55.8 & 36.6 & 12.2 & \textbf{3.2} \\

\random & \textbf{43.5}  && 98.0 & 67.6 & \textbf{14.1} & \textbf{1.8} && 41.8 && 95.5 & 61.7 & 14.8 & 2.5 \\

\bottomrule
\end{tabular}%
}
\caption{
Accuracies for the single-category variants from \S \ref{sec:single_cat}, across per-category subsets of the test set. 
$\mathtt{Full}$ is the original test set (all the categories together). \knowncatshort=\knowncat, \maybeknowncatshort=\maybeknowncat, \weaklyknowncatshort=\weaklyknowncat, \unknowncatshort=\unknowncat. 
In each column, the best result is in \textbf{bold}, as well as the results for which the difference from the best is not statistically significant with $p < 0.05$ (significance test details are in \S \ref{sec:stat_sig_appendix}).
}
\label{tab:categories_analysis}
\end{table*}


\Cref{fig:main_plot} demonstrates that after the development performance peaks at \maxdev (vertical dotted line), it deteriorates as $M$ gradually fits more \unknowncat examples. In this section, we aim to characterize this relationship more accurately by assessing whether a simple linear dependency can tie the impact of fitting \knownSolo and \unknowncat training examples on test accuracy. To this end we use the following linear regression model:
\begin{align}
\text{$Accuracy$} &= \text{$\beta_0$} + \text{$\beta_{\text{kn}}$} \cdot \frac{\text{$N_{\text{kn}}$}}{\text{$|D|$}} + \text{$\beta_{\text{unk}}$} \cdot \frac{\text{$N_{\text{unk}}$}}{\text{$|D|$}}
\label{eq:accuracy}
\end{align}
where $N_\text{Kn}$ and $N_\text{Unk}$ are the number of the \knownSolo and \unknowncat examples in $D$ that $M$ fits.

We estimate the coefficients\footnote{\label{equation_note}Full details in \S \ref{sec:linear_regression_appendix}. We note that this linear model is only valid in bounded region of \text{$N_{\text{kn}} \leq |D|$}, \text{$N_{\text{unk}} \leq |D|$}.} by collecting ($Accuracy$, $N_\text{Kn}$, $N_\text{Unk}$) values after each epoch from models fine-tuned on all $D$ variants. \Cref{tab:linear_model} presents the results (top row).
The high $R^2$ indicates a strong linear relationship between test accuracy and the type of training examples that are fitted.
Our model entails that fitting \unknowncat examples hurts performance ($\beta_{unk} < 0$), while fitting \knownSolo examples improves it ($\beta_\text{kn} > 0$). The estimated negative impact from \unknowncat roughly matches the positive impact from \knownSolo ($|\beta_\text{ukn}| \approx |\beta_\text{kn}|$).

\subsection{Generalization to New Relations}
\label{sec:ood}
In the above setup, the $(q,a)$ pairs in the test set correspond to triplets with the same set of 12 relations appearing in $D$. 
We now investigate whether our observed dynamics has a broader effect on the model's knowledge, and transfers to relations not represented in $D$.
To test this, we reserve a subset of the relations for an \emph{out-of-distribution} (OOD) test set, excluding them from the train and development splits. See \S \ref{sec:data_prep_appendix} for details and Tables \ref{tab:in_domain_test_stats} and
\ref{tab:out_of_domain_test_stats} for in-distribution vs OOD relations.

Our results on the OOD test set reveal similar key insights:
(1) Higher \unknowncat ratio leads to lower OOD test performance and (2) \unknowncat examples are harmful for OOD performance, but mostly when $M$ fits them.
A linear model of the OOD test accuracy (\Cref{eq:accuracy}), shows similar trends: $\beta_\text{unk} < 0$, $\beta_\text{kn} > 0$, $|\beta_\text{ukn}| \approx |\beta_\text{kn}|$ and $R^2=0.95$ (see \Cref{tab:linear_model}). More details are in \S \ref{sec:ood_appendix}.

Overall, \emph{our insights transfer across relations}.
This essentially shows that fine-tuning on \unknowncat examples such as \emph{"Where is [E1] located?"}, can encourage hallucinations on seemingly unrelated questions, such as \emph{"Who founded [E2]?"}. 
This further supports the conclusion that the observed effects likely stem from the model learning the \emph{behavior} of generating answers that are not grounded in its pre-existing knowledge.

    \section{Understanding Knowledge Types: Their Value and Impact}
\label{sec:single_cat}

When addressing our main research question on the effect of \unknowncat fine-tuning examples, we treated
the \knownSolo categories collectively for simplicity (see 
\Cref{tab:categories_def_definitions}).
We now examine the effect of each category, exploring the following questions: \textbf{Q1:} How \emph{training examples} from each category impact the test performance? \textbf{Q2:} What is the model's performance across \emph{test examples} from each category?
To address \textbf{Q1} we created single-category variants of the fine-tuning dataset $D$. A variant of $D$ consisting solely of examples from the category $\mathtt{CAT}$ is denoted as $D_{\mathtt{CAT}}$. 
For reference, we include a variant with the \emph{natural} categories distribution in \eq, denoted \random. $|D|$ is fixed and identical to our experiments in \S \ref{sec:results}. To address \textbf{Q2}, we further break down the test set performance by category. \Cref{tab:categories_analysis} presents the results.

\paragraph{\texttt{MaybeKnown} Examples are Essential.} 
Since \unknowncat examples are harmful, one might expect that it would be best to fine-tune on the most exemplary \knowncat examples. Surprisingly, \known does not obtain the best overall results, as it excels on \knowncat test examples, yet its performance on the remaining categories is inferior. \maybeknown yields the best overall performance. Compared to \known, \maybeknown enhances $M_D$'s performance on \maybeknowncat ($60.1{\mkern -3mu}\rightarrow{\mkern -3mu}69.9$), without compromising performance on \knowncat ($98.7{\mkern -2mu}\rightarrow{\mkern -2mu}98.4$). 
This suggests that \maybeknowncat fine-tuning examples are essential for $M_D$ to correctly handle such examples during inference. 
It also demonstrates that with the right fine-tuning examples, $M_D$ becomes more capable of utilizing its pre-existing knowledge.

\paragraph{Limited Knowledge Enhances Overfitting.}
In \S \ref{sec:harmful_or_neutra}, we demonstrated that \unknowncat fine-tuning examples increase the risk of overfitting. We now observe that this also applies to \weaklyknowncat, though to a lesser degree. Specifically, at \conv,
\weaklyknown and \unknownfull experience significant performance drops compared to \maxdev ($39.2{\mkern -3mu}\rightarrow{\mkern -3mu}35.4$ and $37.5{\mkern -3mu}\rightarrow{\mkern -3mu}25.8$).
With training to \conv, they show a modest improvement on \weaklyknowncat and \unknowncat but substantially degrade on \knowncat and \maybeknowncat.
This highlights that the decrease in performance is strongly attributed to an increased rate of hallucinations w.r.t. facts that were already known to $M$ after pre-training.

Interestingly, \random performs on-par with \maybeknown in \maxdev, suggesting that the mere presence of \maybeknowncat examples in $D$ suffices for high performance on \maybeknowncat, even if $D$ has additional examples from other categories. Yet, \random's performance degrades significantly\footnote{See \S \ref{sec:stat_sig_appendix} for details about this statistic significance test.} after \conv, under-performing \maybeknown\xspace -- indicating that it still suffers from overfitting, most-likely due to the presence of \weaklyknowncat and \unknowncat examples. 
Taken together these results demonstrate  that \maybeknown stands out both in terms of top performance and reduced risk to overfitting.

    \section{\method Knowledge Categories Analysis}
\label{sec:taxonomy}

Assessing a model's knowledge remains an open problem, particularly since evaluating the quality of such methods is challenging due to the lack of ground truth about what the model truly knows.
In this work we proposed \method (\S \ref{sec:categorizing}): a four-category classification of facts w.r.t. the model's knowledge. We now further analyze and discuss our design choices, hoping that \method can serve as a useful taxonomy to guide future research on this subject. 

\paragraph{Fine-grained Known Categories}
We first reflect on whether our choice of splitting \knownSolo into more fine-grained categories, based on the greedy decoding outcome, has been proven meaningful.
As shown in \Cref{tab:categories_analysis}, \knowncat indeed captures facts with high degree of knowledge, as it consistently exceeds $95\%$ accuracy post fine-tuning, while \maybeknowncat and \weaklyknowncat seem to represent weaker knowledge degrees. 
As intended, the performance on \weaklyknowncat is worse that on \maybeknowncat but better than on \unknowncat.
Additionally, the \emph{exact} categories distinction we made was proven useful since it revealed important insights on the importance of the \maybeknowncat fine-tuning examples, as discussed in detail in \S \ref{sec:single_cat}.

\begin{figure}[t]
 \centering
  \vspace{-0.1cm}
 \includegraphics[width=\columnwidth]{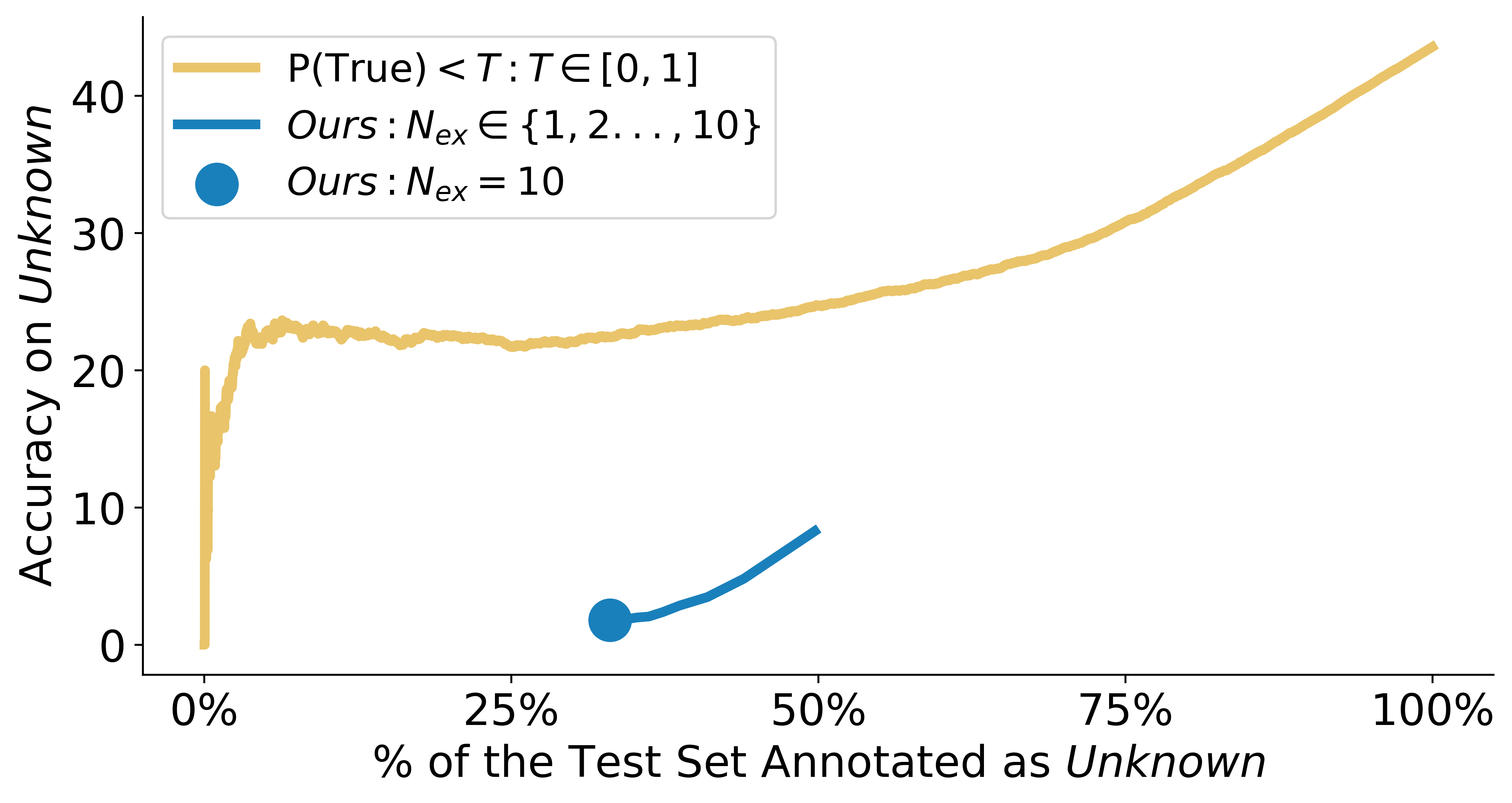}   
    \caption{
    \method \unknowncat categorization vs. classifying examples with P(True) $<T$ as \unknowncat. 
    The x-axis is the $\%$ of test examples classified as \unknowncat and the y-axis is the accuracy on these examples post fine-tuning.
    The \textcolor{ptruecolor}{\textbf{yellow line}} is P(True) for $T\in[0,1]$. Our \unknowncat category is the \textcolor{ourscolor}{\textbf{blue circle}} and the \textcolor{ourscolor}{\textbf{blue line}} corresponds to approximating $\score$ with less than $10$ random 4-shot exemplars (see \S \ref{sec:categorizing} and \S \ref{sec:score_measure_appendix}).
    }
    \label{fig:ptrue_eval}
\end{figure}

\paragraph{Benchmarking Unknown Test Examples}
A desired property for $(q,a)$ pairs classified as \unknowncat that appear in the test set, is that $M$ will incorrectly answer $q$ post fine-tuning (otherwise they are not truly \unknowncat).\footnote{
Since in our closed-book QA setup the train and test sets are disjoint, the model has to rely on its pre-existing knowledge to answer test questions.} In \Cref{tab:categories_analysis} we can see that the accuracy on \unknowncat is extremely low ($3.2\%$ or less), which is a strong indicator that most of the \unknowncat examples are actually unknown to $M$. 

As a case study for comparison, we analyze the P(True) approach by \citet{p_true}: a continuous score that estimates the probability a model assigns to the correctness of a specific answer. 
P(True) was originally used for \emph{self-evaluating} model-generated answers, while we use it to assess whether $M$ considers the ground-truth answer as correct.\footnote{Given a $(q,a)$ pair, P(True) reflects the probability that the model assigns for $a$ being the answer to $q$. It can be applied on a \emph{model-generated} answer for self-evaluation as was done in the original study, or on the \emph{ground-truth} answer to check whether $M$ considers it as correct, like done in our study.}
In \Cref{fig:ptrue_eval}, we explore classifying examples below a P(True) threshold as \unknowncat and compare this methodology to \method.
Our results indicate that, at least in our setting, our approach categorizes \unknowncat examples for which the model's performance after fine-tuning is significantly worse. Specifically, looking at fixed values on the x-axis shows that if we would label a similar fraction of test examples as \unknowncat using both methods, the accuracy on the P(True)-based \unknowncat examples would be much higher post fine-tuning.\footnote{This is a preliminary analysis, and we leave a comprehensive comparison for future work. More details in \S \ref{sec:p_true_appendix}.}
Lastly, the \textcolor{ourscolor}{\textbf{blue line}} shows that using samples from multiple few-shot prompts to approximate $\score$ is crucial, as using $N_{\text{ex}}<10$ leads to higher test accuracy on \method \unknowncat examples.
    \section{Fine-tuning to Abstain on \unknowncat Examples}
\label{sec:idk_appendix}

We showed that fitting \unknowncat fine-tuning examples negatively affects the test performance (\S \ref{sec:harmful_or_neutra} and \S \ref{sec:equation}). However, this negative effect manifests as a form of \emph{overfitting}.
From practical perspective, we showed that we can mitigate overfitting by either using early-stopping or filtering-out \unknowncat examples from the fine-tuning dataset. 

We now explore an additional approach where we fine-tune the model to abstain from \unknowncat examples as a potential mitigation. Specifically, we replace the label of the \unknowncat fine-tuning examples with the expression \emph{``I don't know''} and test whether this mitigates the observed overfitting. 

\Cref{tab:idk} presents the $\%$ of the test questions that were answered (i.e. $M_D$ did not respond with \emph{``I don't know''}) and the accuracy on those questions.
Consistent with the findings from previous work \cite{zhang2023r}, we observe an improved accuracy on willingly answered test examples (when comparing $D$ vs \didk). 
When we compare \maxdev vs \conv for $D$ we observe a performance drop ($43.0 \rightarrow 38.8$) which illustrates the overfitting effect. However, we observe that re-labeling the \unknowncat examples with uncertainty expression seem to reduce the risk of overfitting. Specifically, the accuracy for \didk remains $61.8$ for both \maxdev and \conv, with a small decrease on the number of willingly answered questions ($58.7 \rightarrow 55.6$).

\section{Discussion}
\label{sec:dicsussion}

\paragraph{Practical Implications.}
This work highlights the risk in using supervised fine-tuning to update LLMs’ knowledge, as we present empirical evidence that acquiring new knowledge through fine-tuning is correlated with hallucinations w.r.t pre-existing knowledge.
Additionally, this work raises important questions for future exploration regarding fine-tuning practices.
We saw that \unknowncat examples are fitted slower than the \knownSolo ones, thus their negative effect manifests as a form of \emph{overfitting}, which emphasizes the importance of using \emph{early-stopping} instead of a fixed number of fine-tuning steps.
However, early-stopping may be less effective when fine-tuning on numerous tasks with distinct optimal stopping points.
An alternative solution can be to 
avoid adding new knowledge, by aligning
the fine-tuning data with the model's knowledge through filtering-out \unknowncat examples. 
We show initial evidence that this can reduce the risk of overfitting without compromising performance.
\looseness=-1
A possible drawback of filtering, is that \unknowncat fine-tuning examples can still be useful to teach LLMs to express uncertainty on \unknowncat test examples \cite{zhang2023r,yang2023alignment}.
This raises the question: \emph{can re-labeling \unknowncat fine-tuning examples with uncertainty expressions} (e.g., \emph{``I don't know''}) \emph{reduce their negative effect?}
Our experiment (\S \ref{sec:idk_appendix}) suggests that the answer is \emph{yes}, which indicates that such approaches could be the most promising.

\begin{table}[t]
\centering
\resizebox{\columnwidth}{!}{%
\begin{tabular}{lrrcrr}

\multicolumn{1}{c}{\multirow{1}{*}{}} &
\multicolumn{2}{c}{\maxdev} &
\phantom{abc} &
\multicolumn{2}{c}{\conv} \\

\cmidrule{2-3} \cmidrule{5-6} 

\multicolumn{1}{c}{\multirow{1}{*}{}} &
  \multicolumn{1}{c}{Accuracy} &
  \multicolumn{1}{c}{\% Answered} &&
  \multicolumn{1}{c}{Accuracy} &
  \multicolumn{1}{c}{\% Answered} \\
\toprule

$D$ & 43.0 & 100.0 &&  38.8 & 100.0 \\
\didk & 61.8 & 58.7 &&  61.8 & 55.6 \\

\bottomrule
\end{tabular}%
}
\caption{
Results where the label of the \unknowncat fine-tuning examples is replaced with \emph{``I don't know''}. $D$ in this case is the variant with $50\%$ \knownSolo and $50\%$ \unknowncat. \didk is the variant where all the $50\%$ \unknowncat fine-tuning examples were re-labeled with \emph{``I don't know''}.
The accuracy is measured on the subset of the test questions that were answered, i.e. $M_D$ did not respond with \emph{``I don't know''}.
}
\label{tab:idk}
\end{table}


\paragraph{Superficial Alignment Hypothesis.}
\citet{LIMA} hypothesized that the knowledge and capabilities of LLMs are mostly learned during pre-training, 
while alignment is a simple process where the model learns the style or format for interacting with users. They substantiate this hypothesis by showing that fine-tuning on just $\mathtt{1k}$ high-quality examples can result with a competitive assistant LLM, named LIMA. As discussed in \S \ref{sec:ignore}, we show evidence that LLMs struggle to acquire new knowledge present in the \unknowncat examples and mostly learn to utilize their pre-existing knowledge. We also showed that fine-tuning on \knowncat examples led to sub-optimal utilization of pre-existing knowledge, despite our task format being simpler than LIMA's and our dataset being six times larger. Taken together, our findings suggest that even though most of the LLM's knowledge is indeed acquired through pre-training, the model learns more than just style or format through fine-tuning, as the selection of fine-tuning examples significantly influences the model’s capability to utilize its pre-existing knowledge post fine-tuning.

    \section{Related Work}
\label{sec:related}

\paragraph{New knowledge and hallucinations.}
\citet{Schulman_RL}, \citet{Goldberg_RL} and \citet{The_False_Promise_of_Imitating_LLMs} mention the conjecture that fine-tuning on new factual knowledge may encourage hallucinations.
\citet{hallucinations_survey} categorized hallucination causes and formally defined this scenario as \emph{capability misalignment}, also highlighting that limited research addresses capability misalignment due to the challenge of defining the knowledge boundary of LLMs.

Recent work support our findings. For instance,
\citet{DBLP:conf/icml/GhosalHR24} showed that models fine-tuned on well-known facts exhibit enhanced factuality compared to those fine-tuned on unpopular facts, which can be attributed to the model’s lesser familiarity with unpopular facts. 
Another example is \citet{lin2024flame}, who fine-tuned a model using data generated by either a pre-trained model or a retrieval-augmented variant. They found that the latter resulted in reduced factuality, which can be attributed to the introduction of new factual knowledge in the retrieved texts.
\citet{ren-etal-2024-learning} have also investigated the effects of introducing new factual knowledge through fine-tuning in a considerably different methodological setup, focusing on multiple-choice questions, conducting relatively short fine-tuning runs, and testing only $100\%$ known and $100\%$ unknown mixtures. Their results align with ours, which further reinforces our conclusions.
Lastly, these insights were also integrated into the instruction-tuning phase of Llama 3 models \cite{dubey2024llama}, ensuring that the examples are aligned with pre-training knowledge.

Another line of work explores the model's behavior on new knowledge in test time. \citet{Unfamiliar_SFT} showed that when a fine-tuned LLM encounters unknown queries at test time, its responses mimic the responses associated with the unknown examples in the fine-tuning data. \citet{ALCUNA} showed that LLMs' performance is not satisfactory when they face new knowledge in their input contexts and
\citet{Crafting_In_context_Examples} analyzed the impact of unknown \emph{in-context} learning examples.


\paragraph{Quantifying knowledge in LLMs.}
\method 
can be seen as a 
confidence elicitation method for the ground truth label ($M$ \emph{knows} $(q,a)$ if it is confident that $a$ is the answer to $q$).
Existing work derive calibrated confidence from LLMs by
examining agreement across multiple samples \cite{kuhn2023semantic, manakul2023selfcheckgpt, tian2023fine,lyu2024calibrating}, probing internal representations
\cite{azaria2023internal, burns2022discovering},
eliciting verbalized probability \cite{tian2023just} or direct prompting \cite{p_true}.
\citeauthor{p_true} also trained a separate P(IK) model to predict if the LLM knows the answer to $q$. The label for P(IK) was approximated by the fraction of correct sampled answers, which is conceptually aligned with $\score$ (\S \ref{sec:categorizing}).
A key difference is that we also define the \method categories, and provide evidence that we capture meaningful and useful categories.

    \section{Conclusion}
\label{sec:conclusion}

We study the impact of integrating new factual knowledge through fine-tuning on the model's tendency to hallucinate. 
We first propose \method, 
a categorization of facts w.r.t. LLM's knowledge.
We then design a controlled study where we isolate the impact of new knowledge and rigorously evaluate its effects. 
We provide multiple insights on the fine-tuning dynamics, with the following key findings: (1) Acquiring new knowledge via supervised fine-tuning is correlated with hallucinations w.r.t. pre-existing knowledge.
(2) LLMs struggle to integrate new knowledge through fine-tuning and mostly learn to use their pre-existing knowledge.

\section{Limitations}
\label{sec:limitations}

Our experiments were conducted using a single LLM, and thus it is unclear whether results will vary with different LLMs. Having said that, our study is extremely compute-heavy and thus challenging to replicate on multiple LLMs: First, our fine-tuning is compute-heavy as its runs are very long as we wanted to analyze the behavior during different stages of fine-tuning (including the overfitting stages). Second, and most importantly, to facilitate our study we needed to annotate a large scale dataset w.r.t the \method categories.
To derive reliable conclusions, it was crucial to accurately assess the model's knowledge w.r.t. a single fine-tuning example. In our case we run 170 inference steps per example, i.e., more than $15M$ inference steps to categorize our full dataset.

In addition, since we focus on closed-book QA, the practical implications from our study such as filtering-out \unknowncat fine-tuning examples still require validation in settings involving long-form text generation.
To filter-out examples that introduce new factual knowledge in long-form generation tasks, one would need to make adaptations to \method and come up with an effective way to compare the sampled answer with the ground-truth to approximate $\score$. We leave this for future work. Long-form generation tasks introduce evaluation challenges, leading to a wide adoption of LLM-based evaluations. Our choice to focus explicitly on closed book QA facilitates more accurate evaluation that enhances the reliability of our findings.

Lastly, we did not test the effect of adding additional fine-tuning examples from diverse tasks into the fine-tuning mixture.
While this could more closely approximate a typical instruction fine-tuning scenario, such dataset extension may introduce new factual knowledge in an uncontrollable way, which will limit our findings.

\section{Acknowledgments}
\label{sec:ack}
We thank the reviewers for their valuable comments
and suggestions for improving this work. We would also like to thank Ori Ram, Uri Shaham, Alon Jacovi, Mor Ventura, Yochai Blau, Eyal Ben-David, Avi Caciularu, Avinatan Hassidim and the members of Roi Reichart's NLP group for reviewing the paper draft and providing valuable feedback. Special thanks to Uri Shaham for assisting in setting up the fine-tuning pipeline during the early stages of our research.
    \bibliography{custom}

\begin{thebibliography}{58}
\expandafter\ifx\csname natexlab\endcsname\relax\def\natexlab#1{#1}\fi

\bibitem[{AlKhamissi et~al.(2022)AlKhamissi, Li, Celikyilmaz, Diab, and
  Ghazvininejad}]{LLMs_as_KG_4}
Badr AlKhamissi, Millicent Li, Asli Celikyilmaz, Mona Diab, and Marjan
  Ghazvininejad. 2022.
\newblock A review on language models as knowledge bases.
\newblock \emph{arXiv preprint arXiv:2204.06031}.

\bibitem[{Anil et~al.(2023)Anil, Dai, Firat, Johnson, Lepikhin, Passos,
  Shakeri, Taropa, Bailey, Chen et~al.}]{PaLM2}
Rohan Anil, Andrew~M Dai, Orhan Firat, Melvin Johnson, Dmitry Lepikhin,
  Alexandre Passos, Siamak Shakeri, Emanuel Taropa, Paige Bailey, Zhifeng Chen,
  et~al. 2023.
\newblock Palm 2 technical report.
\newblock \emph{arXiv preprint arXiv:2305.10403}.

\bibitem[{Azaria and Mitchell(2023)}]{azaria2023internal}
Amos Azaria and Tom Mitchell. 2023.
\newblock The internal state of an llm knows when its lying.
\newblock \emph{arXiv preprint arXiv:2304.13734}.

\bibitem[{Biderman et~al.(2024)Biderman, Portes, Ortiz, Paul, Greengard,
  Jennings, King, Havens, Chiley, Frankle, Blakeney, and
  Cunningham}]{biderman2024lora}
Dan Biderman, Jacob Portes, Jose Javier~Gonzalez Ortiz, Mansheej Paul, Philip
  Greengard, Connor Jennings, Daniel King, Sam Havens, Vitaliy Chiley, Jonathan
  Frankle, Cody Blakeney, and John~Patrick Cunningham. 2024.
\newblock \href {https://openreview.net/forum?id=aloEru2qCG} {Lo{RA} learns
  less and forgets less}.
\newblock \emph{Transactions on Machine Learning Research}.
\newblock Featured Certification.

\bibitem[{Burns et~al.(2022)Burns, Ye, Klein, and
  Steinhardt}]{burns2022discovering}
Collin Burns, Haotian Ye, Dan Klein, and Jacob Steinhardt. 2022.
\newblock Discovering latent knowledge in language models without supervision.
\newblock \emph{arXiv preprint arXiv:2212.03827}.

\bibitem[{Chern et~al.(2023)Chern, Chern, Chen, Yuan, Feng, Zhou, He, Neubig,
  Liu et~al.}]{chern2023factool}
I~Chern, Steffi Chern, Shiqi Chen, Weizhe Yuan, Kehua Feng, Chunting Zhou,
  Junxian He, Graham Neubig, Pengfei Liu, et~al. 2023.
\newblock Factool: Factuality detection in generative ai--a tool augmented
  framework for multi-task and multi-domain scenarios.
\newblock \emph{arXiv preprint arXiv:2307.13528}.

\bibitem[{Cohen et~al.(2023)Cohen, Geva, Berant, and Globerson}]{LLMs_as_KG_3}
Roi Cohen, Mor Geva, Jonathan Berant, and Amir Globerson. 2023.
\newblock \href {https://doi.org/10.18653/v1/2023.findings-eacl.139} {Crawling
  the internal knowledge-base of language models}.
\newblock In \emph{Findings of the Association for Computational Linguistics:
  EACL 2023}, pages 1856--1869, Dubrovnik, Croatia. Association for
  Computational Linguistics.

\bibitem[{Dubey et~al.(2024)Dubey, Jauhri, Pandey, Kadian, Al-Dahle, Letman,
  Mathur, Schelten, Yang, Fan et~al.}]{dubey2024llama}
Abhimanyu Dubey, Abhinav Jauhri, Abhinav Pandey, Abhishek Kadian, Ahmad
  Al-Dahle, Aiesha Letman, Akhil Mathur, Alan Schelten, Amy Yang, Angela Fan,
  et~al. 2024.
\newblock The llama 3 herd of models.
\newblock \emph{arXiv preprint arXiv:2407.21783}.

\bibitem[{Gao(2021)}]{Behavior_Cloning_is_Miscalibrated}
Leo Gao. 2021.
\newblock \href
  {https://www.alignmentforum.org/posts/BgoKdAzogxmgkuuAt/behavior-cloning-is-miscalibrated}
  {Behavior cloning is miscalibrated}.
\newblock \emph{AI Alignment Forum}.

\bibitem[{Gekhman et~al.(2023)Gekhman, Herzig, Aharoni, Elkind, and
  Szpektor}]{gekhman-etal-2023-trueteacher}
Zorik Gekhman, Jonathan Herzig, Roee Aharoni, Chen Elkind, and Idan Szpektor.
  2023.
\newblock \href {https://doi.org/10.18653/v1/2023.emnlp-main.127}
  {{T}rue{T}eacher: Learning factual consistency evaluation with large language
  models}.
\newblock In \emph{Proceedings of the 2023 Conference on Empirical Methods in
  Natural Language Processing}, pages 2053--2070, Singapore. Association for
  Computational Linguistics.

\bibitem[{Ghosal et~al.(2024)Ghosal, Hashimoto, and
  Raghunathan}]{DBLP:conf/icml/GhosalHR24}
Gaurav~Rohit Ghosal, Tatsunori Hashimoto, and Aditi Raghunathan. 2024.
\newblock \href {https://openreview.net/forum?id=cPsn9AcOYh} {Understanding
  finetuning for factual knowledge extraction}.
\newblock In \emph{Forty-first International Conference on Machine Learning,
  {ICML} 2024, Vienna, Austria, July 21-27, 2024}. OpenReview.net.

\bibitem[{Goldberg(2023)}]{Goldberg_RL}
Yoav Goldberg. 2023.
\newblock \href
  {https://gist.github.com/yoavg/6bff0fecd65950898eba1bb321cfbd81}
  {Reinforcement learning for language models}.

\bibitem[{Gudibande et~al.(2023)Gudibande, Wallace, Snell, Geng, Liu, Abbeel,
  Levine, and Song}]{The_False_Promise_of_Imitating_LLMs}
Arnav Gudibande, Eric Wallace, Charlie Snell, Xinyang Geng, Hao Liu, Pieter
  Abbeel, Sergey Levine, and Dawn Song. 2023.
\newblock The false promise of imitating proprietary llms.
\newblock \emph{arXiv preprint arXiv:2305.15717}.

\bibitem[{Han et~al.(2024)Han, Gao, Liu, Zhang et~al.}]{han2024parameter}
Zeyu Han, Chao Gao, Jinyang Liu, Sai~Qian Zhang, et~al. 2024.
\newblock Parameter-efficient fine-tuning for large models: A comprehensive
  survey.
\newblock \emph{arXiv preprint arXiv:2403.14608}.

\bibitem[{Honovich et~al.(2022)Honovich, Aharoni, Herzig, Taitelbaum,
  Kukliansy, Cohen, Scialom, Szpektor, Hassidim, and
  Matias}]{honovich-etal-2022-true-evaluating}
Or~Honovich, Roee Aharoni, Jonathan Herzig, Hagai Taitelbaum, Doron Kukliansy,
  Vered Cohen, Thomas Scialom, Idan Szpektor, Avinatan Hassidim, and Yossi
  Matias. 2022.
\newblock \href {https://doi.org/10.18653/v1/2022.naacl-main.287} {{TRUE}:
  Re-evaluating factual consistency evaluation}.
\newblock In \emph{Proceedings of the 2022 Conference of the North American
  Chapter of the Association for Computational Linguistics: Human Language
  Technologies}, pages 3905--3920, Seattle, United States. Association for
  Computational Linguistics.

\bibitem[{Honovich et~al.(2021)Honovich, Choshen, Aharoni, Neeman, Szpektor,
  and Abend}]{honovich-etal-2021-q2}
Or~Honovich, Leshem Choshen, Roee Aharoni, Ella Neeman, Idan Szpektor, and Omri
  Abend. 2021.
\newblock \href {https://doi.org/10.18653/v1/2021.emnlp-main.619} {$q^{2}$:
  {E}valuating factual consistency in knowledge-grounded dialogues via question
  generation and question answering}.
\newblock In \emph{Proceedings of the 2021 Conference on Empirical Methods in
  Natural Language Processing}, pages 7856--7870, Online and Punta Cana,
  Dominican Republic. Association for Computational Linguistics.

\bibitem[{Hu et~al.(2022)Hu, Shen, Wallis, Allen{-}Zhu, Li, Wang, Wang, and
  Chen}]{DBLP:conf/iclr/HuSWALWWC22}
Edward~J. Hu, Yelong Shen, Phillip Wallis, Zeyuan Allen{-}Zhu, Yuanzhi Li,
  Shean Wang, Lu~Wang, and Weizhu Chen. 2022.
\newblock \href {https://openreview.net/forum?id=nZeVKeeFYf9} {Lora: Low-rank
  adaptation of large language models}.
\newblock In \emph{The Tenth International Conference on Learning
  Representations, {ICLR} 2022, Virtual Event, April 25-29, 2022}.
  OpenReview.net.

\bibitem[{Huang et~al.(2023)Huang, Yu, Ma, Zhong, Feng, Wang, Chen, Peng, Feng,
  Qin et~al.}]{hallucinations_survey}
Lei Huang, Weijiang Yu, Weitao Ma, Weihong Zhong, Zhangyin Feng, Haotian Wang,
  Qianglong Chen, Weihua Peng, Xiaocheng Feng, Bing Qin, et~al. 2023.
\newblock A survey on hallucination in large language models: Principles,
  taxonomy, challenges, and open questions.
\newblock \emph{arXiv preprint arXiv:2311.05232}.

\bibitem[{Ibrahim et~al.(2024)Ibrahim, Th{\'e}rien, Gupta, Richter, Anthony,
  Belilovsky, Lesort, and Rish}]{ibrahim2024simple}
Adam Ibrahim, Benjamin Th{\'e}rien, Kshitij Gupta, Mats~Leon Richter,
  Quentin~Gregory Anthony, Eugene Belilovsky, Timoth{\'e}e Lesort, and Irina
  Rish. 2024.
\newblock \href {https://openreview.net/forum?id=DimPeeCxKO} {Simple and
  scalable strategies to continually pre-train large language models}.
\newblock \emph{Transactions on Machine Learning Research}.

\bibitem[{Jiang et~al.(2024)Jiang, Sun, Shi, Rodriguez, Zhou, Neubig, Lin, Yih,
  and Iyer}]{jiang-etal-2024-instruction}
Zhengbao Jiang, Zhiqing Sun, Weijia Shi, Pedro Rodriguez, Chunting Zhou, Graham
  Neubig, Xi~Lin, Wen-tau Yih, and Srini Iyer. 2024.
\newblock \href {https://aclanthology.org/2024.acl-long.296} {Instruction-tuned
  language models are better knowledge learners}.
\newblock In \emph{Proceedings of the 62nd Annual Meeting of the Association
  for Computational Linguistics (Volume 1: Long Papers)}, pages 5421--5434,
  Bangkok, Thailand. Association for Computational Linguistics.

\bibitem[{Kadavath et~al.(2022)Kadavath, Conerly, Askell, Henighan, Drain,
  Perez, Schiefer, Hatfield-Dodds, DasSarma, Tran-Johnson et~al.}]{p_true}
Saurav Kadavath, Tom Conerly, Amanda Askell, Tom Henighan, Dawn Drain, Ethan
  Perez, Nicholas Schiefer, Zac Hatfield-Dodds, Nova DasSarma, Eli
  Tran-Johnson, et~al. 2022.
\newblock Language models (mostly) know what they know.
\newblock \emph{arXiv preprint arXiv:2207.05221}.

\bibitem[{Kamalloo et~al.(2023)Kamalloo, Dziri, Clarke, and Rafiei}]{EM_2}
Ehsan Kamalloo, Nouha Dziri, Charles L.~A. Clarke, and Davood Rafiei. 2023.
\newblock \href {https://doi.org/10.18653/V1/2023.ACL-LONG.307} {Evaluating
  open-domain question answering in the era of large language models}.
\newblock In \emph{Proceedings of the 61st Annual Meeting of the Association
  for Computational Linguistics (Volume 1: Long Papers), {ACL} 2023, Toronto,
  Canada, July 9-14, 2023}, pages 5591--5606. Association for Computational
  Linguistics.

\bibitem[{Kang et~al.(2024)Kang, Wallace, Tomlin, Kumar, and
  Levine}]{Unfamiliar_SFT}
Katie Kang, Eric Wallace, Claire Tomlin, Aviral Kumar, and Sergey Levine. 2024.
\newblock Unfamiliar finetuning examples control how language models
  hallucinate.
\newblock \emph{arXiv preprint arXiv:2403.05612}.

\bibitem[{Kryscinski et~al.(2020)Kryscinski, McCann, Xiong, and
  Socher}]{kryscinski-etal-2020-evaluating}
Wojciech Kryscinski, Bryan McCann, Caiming Xiong, and Richard Socher. 2020.
\newblock \href {https://doi.org/10.18653/v1/2020.emnlp-main.750} {Evaluating
  the factual consistency of abstractive text summarization}.
\newblock In \emph{Proceedings of the 2020 Conference on Empirical Methods in
  Natural Language Processing (EMNLP)}, pages 9332--9346, Online. Association
  for Computational Linguistics.

\bibitem[{Kuhn et~al.(2023)Kuhn, Gal, and Farquhar}]{kuhn2023semantic}
Lorenz Kuhn, Yarin Gal, and Sebastian Farquhar. 2023.
\newblock Semantic uncertainty: Linguistic invariances for uncertainty
  estimation in natural language generation.
\newblock \emph{arXiv preprint arXiv:2302.09664}.

\bibitem[{Laban et~al.(2022)Laban, Schnabel, Bennett, and
  Hearst}]{laban-etal-2022-summac}
Philippe Laban, Tobias Schnabel, Paul~N. Bennett, and Marti~A. Hearst. 2022.
\newblock \href {https://doi.org/10.1162/tacl_a_00453} {{S}umma{C}: Re-visiting
  {NLI}-based models for inconsistency detection in summarization}.
\newblock \emph{Transactions of the Association for Computational Linguistics},
  10:163--177.

\bibitem[{Lee et~al.(2023)Lee, Atreya, Ye, and
  Choi}]{Crafting_In_context_Examples}
Yoonsang Lee, Pranav Atreya, Xi~Ye, and Eunsol Choi. 2023.
\newblock Crafting in-context examples according to lms' parametric knowledge.
\newblock \emph{arXiv preprint arXiv:2311.09579}.

\bibitem[{Lin et~al.(2023)Lin, Ravichander, Lu, Dziri, Sclar, Chandu,
  Bhagavatula, and Choi}]{The_Unlocking_Spell_on_Base_LLMs}
Bill~Yuchen Lin, Abhilasha Ravichander, Ximing Lu, Nouha Dziri, Melanie Sclar,
  Khyathi Chandu, Chandra Bhagavatula, and Yejin Choi. 2023.
\newblock \href {http://arxiv.org/abs/2312.01552} {The unlocking spell on base
  llms: Rethinking alignment via in-context learning}.
\newblock \emph{ArXiv preprint}.

\bibitem[{Lin et~al.(2024)Lin, Gao, Oguz, Xiong, Lin, Yih, and
  Chen}]{lin2024flame}
Sheng-Chieh Lin, Luyu Gao, Barlas Oguz, Wenhan Xiong, Jimmy Lin, Wen-tau Yih,
  and Xilun Chen. 2024.
\newblock Flame: Factuality-aware alignment for large language models.
\newblock \emph{arXiv preprint arXiv:2405.01525}.

\bibitem[{Lyu et~al.(2024)Lyu, Shridhar, Malaviya, Zhang, Elazar, Tandon,
  Apidianaki, Sachan, and Callison-Burch}]{lyu2024calibrating}
Qing Lyu, Kumar Shridhar, Chaitanya Malaviya, Li~Zhang, Yanai Elazar, Niket
  Tandon, Marianna Apidianaki, Mrinmaya Sachan, and Chris Callison-Burch. 2024.
\newblock Calibrating large language models with sample consistency.
\newblock \emph{arXiv preprint arXiv:2402.13904}.

\bibitem[{Manakul et~al.(2023)Manakul, Liusie, and
  Gales}]{manakul2023selfcheckgpt}
Potsawee Manakul, Adian Liusie, and Mark~JF Gales. 2023.
\newblock Selfcheckgpt: Zero-resource black-box hallucination detection for
  generative large language models.
\newblock \emph{arXiv preprint arXiv:2303.08896}.

\bibitem[{Meng et~al.(2022)Meng, Bau, Andonian, and
  Belinkov}]{DBLP:conf/nips/MengBAB22}
Kevin Meng, David Bau, Alex Andonian, and Yonatan Belinkov. 2022.
\newblock \href
  {http://papers.nips.cc/paper\_files/paper/2022/hash/6f1d43d5a82a37e89b0665b33bf3a182-Abstract-Conference.html}
  {Locating and editing factual associations in {GPT}}.
\newblock In \emph{Advances in Neural Information Processing Systems 35: Annual
  Conference on Neural Information Processing Systems 2022, NeurIPS 2022, New
  Orleans, LA, USA, November 28 - December 9, 2022}.

\bibitem[{Meng et~al.(2023)Meng, Sharma, Andonian, Belinkov, and
  Bau}]{DBLP:conf/iclr/MengSABB23}
Kevin Meng, Arnab~Sen Sharma, Alex~J. Andonian, Yonatan Belinkov, and David
  Bau. 2023.
\newblock \href {https://openreview.net/forum?id=MkbcAHIYgyS} {Mass-editing
  memory in a transformer}.
\newblock In \emph{The Eleventh International Conference on Learning
  Representations, {ICLR} 2023, Kigali, Rwanda, May 1-5, 2023}. OpenReview.net.

\bibitem[{Min et~al.(2023)Min, Krishna, Lyu, Lewis, Yih, Koh, Iyyer,
  Zettlemoyer, and Hajishirzi}]{min-etal-2023-factscore}
Sewon Min, Kalpesh Krishna, Xinxi Lyu, Mike Lewis, Wen-tau Yih, Pang Koh, Mohit
  Iyyer, Luke Zettlemoyer, and Hannaneh Hajishirzi. 2023.
\newblock \href {https://doi.org/10.18653/v1/2023.emnlp-main.741}
  {{FA}ct{S}core: Fine-grained atomic evaluation of factual precision in long
  form text generation}.
\newblock In \emph{Proceedings of the 2023 Conference on Empirical Methods in
  Natural Language Processing}, pages 12076--12100, Singapore. Association for
  Computational Linguistics.

\bibitem[{Mishra et~al.(2022)Mishra, Khashabi, Baral, and
  Hajishirzi}]{NATURAL_INSTRUCTIONS}
Swaroop Mishra, Daniel Khashabi, Chitta Baral, and Hannaneh Hajishirzi. 2022.
\newblock \href {https://doi.org/10.18653/v1/2022.acl-long.244} {Cross-task
  generalization via natural language crowdsourcing instructions}.
\newblock In \emph{Proceedings of the 60th Annual Meeting of the Association
  for Computational Linguistics (Volume 1: Long Papers)}, pages 3470--3487,
  Dublin, Ireland. Association for Computational Linguistics.

\bibitem[{Ouyang et~al.(2022)Ouyang, Wu, Jiang, Almeida, Wainwright, Mishkin,
  Zhang, Agarwal, Slama, Ray, Schulman, Hilton, Kelton, Miller, Simens, Askell,
  Welinder, Christiano, Leike, and Lowe}]{RL_OpenAI}
Long Ouyang, Jeffrey Wu, Xu~Jiang, Diogo Almeida, Carroll~L. Wainwright, Pamela
  Mishkin, Chong Zhang, Sandhini Agarwal, Katarina Slama, Alex Ray, John
  Schulman, Jacob Hilton, Fraser Kelton, Luke Miller, Maddie Simens, Amanda
  Askell, Peter Welinder, Paul~F. Christiano, Jan Leike, and Ryan Lowe. 2022.
\newblock \href
  {http://papers.nips.cc/paper\_files/paper/2022/hash/b1efde53be364a73914f58805a001731-Abstract-Conference.html}
  {Training language models to follow instructions with human feedback}.
\newblock In \emph{Advances in Neural Information Processing Systems 35: Annual
  Conference on Neural Information Processing Systems 2022, NeurIPS 2022, New
  Orleans, LA, USA, November 28 - December 9, 2022}.

\bibitem[{Parmar et~al.(2024)Parmar, Satheesh, Patwary, Shoeybi, and
  Catanzaro}]{parmar2024reuse}
Jupinder Parmar, Sanjev Satheesh, Mostofa Patwary, Mohammad Shoeybi, and Bryan
  Catanzaro. 2024.
\newblock Reuse, don't retrain: A recipe for continued pretraining of language
  models.
\newblock \emph{arXiv preprint arXiv:2407.07263}.

\bibitem[{Petroni et~al.(2019)Petroni, Rockt{\"a}schel, Riedel, Lewis, Bakhtin,
  Wu, and Miller}]{LLMs_as_KG_1}
Fabio Petroni, Tim Rockt{\"a}schel, Sebastian Riedel, Patrick Lewis, Anton
  Bakhtin, Yuxiang Wu, and Alexander Miller. 2019.
\newblock \href {https://doi.org/10.18653/v1/D19-1250} {Language models as
  knowledge bases?}
\newblock In \emph{Proceedings of the 2019 Conference on Empirical Methods in
  Natural Language Processing and the 9th International Joint Conference on
  Natural Language Processing (EMNLP-IJCNLP)}, pages 2463--2473, Hong Kong,
  China. Association for Computational Linguistics.

\bibitem[{Rafailov et~al.(2024)Rafailov, Sharma, Mitchell, Manning, Ermon, and
  Finn}]{rafailov2024direct}
Rafael Rafailov, Archit Sharma, Eric Mitchell, Christopher~D Manning, Stefano
  Ermon, and Chelsea Finn. 2024.
\newblock Direct preference optimization: Your language model is secretly a
  reward model.
\newblock \emph{Advances in Neural Information Processing Systems}, 36.

\bibitem[{Rajpurkar et~al.(2016)Rajpurkar, Zhang, Lopyrev, and Liang}]{SQUAD}
Pranav Rajpurkar, Jian Zhang, Konstantin Lopyrev, and Percy Liang. 2016.
\newblock \href {https://doi.org/10.18653/v1/D16-1264} {{SQ}u{AD}: 100,000+
  questions for machine comprehension of text}.
\newblock In \emph{Proceedings of the 2016 Conference on Empirical Methods in
  Natural Language Processing}, pages 2383--2392, Austin, Texas. Association
  for Computational Linguistics.

\bibitem[{Ren et~al.(2024)Ren, Cao, Lin, Liu, Han, Zeng, Guanglu, Cai, and
  Sun}]{ren-etal-2024-learning}
Mengjie Ren, Boxi Cao, Hongyu Lin, Cao Liu, Xianpei Han, Ke~Zeng, Wan Guanglu,
  Xunliang Cai, and Le~Sun. 2024.
\newblock \href {https://aclanthology.org/2024.acl-long.330} {Learning or
  self-aligning? rethinking instruction fine-tuning}.
\newblock In \emph{Proceedings of the 62nd Annual Meeting of the Association
  for Computational Linguistics (Volume 1: Long Papers)}, pages 6090--6105,
  Bangkok, Thailand. Association for Computational Linguistics.

\bibitem[{Rubin et~al.(2022)Rubin, Herzig, and
  Berant}]{Learning_To_Retrieve_Prompts}
Ohad Rubin, Jonathan Herzig, and Jonathan Berant. 2022.
\newblock \href {https://doi.org/10.18653/v1/2022.naacl-main.191} {Learning to
  retrieve prompts for in-context learning}.
\newblock In \emph{Proceedings of the 2022 Conference of the North American
  Chapter of the Association for Computational Linguistics: Human Language
  Technologies}, pages 2655--2671, Seattle, United States. Association for
  Computational Linguistics.

\bibitem[{Schulman(2023)}]{Schulman_RL}
John Schulman. 2023.
\newblock \href
  {https://www.youtube.com/watch?v=hhiLw5Q_UFg&ab_channel=BerkeleyEECS}
  {Reinforcement learning from human feedback: Progress and challenges}.

\bibitem[{Scialom et~al.(2021)Scialom, Dray, Lamprier, Piwowarski, Staiano,
  Wang, and Gallinari}]{scialom-etal-2021-questeval}
Thomas Scialom, Paul-Alexis Dray, Sylvain Lamprier, Benjamin Piwowarski, Jacopo
  Staiano, Alex Wang, and Patrick Gallinari. 2021.
\newblock \href {https://doi.org/10.18653/v1/2021.emnlp-main.529}
  {{Q}uest{E}val: Summarization asks for fact-based evaluation}.
\newblock In \emph{Proceedings of the 2021 Conference on Empirical Methods in
  Natural Language Processing}, pages 6594--6604, Online and Punta Cana,
  Dominican Republic. Association for Computational Linguistics.

\bibitem[{Sciavolino et~al.(2021)Sciavolino, Zhong, Lee, and
  Chen}]{Entity_Questions}
Christopher Sciavolino, Zexuan Zhong, Jinhyuk Lee, and Danqi Chen. 2021.
\newblock \href {https://doi.org/10.18653/V1/2021.EMNLP-MAIN.496} {Simple
  entity-centric questions challenge dense retrievers}.
\newblock In \emph{Proceedings of the 2021 Conference on Empirical Methods in
  Natural Language Processing, {EMNLP} 2021, Virtual Event / Punta Cana,
  Dominican Republic, 7-11 November, 2021}, pages 6138--6148. Association for
  Computational Linguistics.

\bibitem[{Tian et~al.(2023{\natexlab{a}})Tian, Mitchell, Yao, Manning, and
  Finn}]{tian2023fine}
Katherine Tian, Eric Mitchell, Huaxiu Yao, Christopher~D Manning, and Chelsea
  Finn. 2023{\natexlab{a}}.
\newblock Fine-tuning language models for factuality.
\newblock \emph{arXiv preprint arXiv:2311.08401}.

\bibitem[{Tian et~al.(2023{\natexlab{b}})Tian, Mitchell, Zhou, Sharma,
  Rafailov, Yao, Finn, and Manning}]{tian2023just}
Katherine Tian, Eric Mitchell, Allan Zhou, Archit Sharma, Rafael Rafailov,
  Huaxiu Yao, Chelsea Finn, and Christopher~D Manning. 2023{\natexlab{b}}.
\newblock Just ask for calibration: Strategies for eliciting calibrated
  confidence scores from language models fine-tuned with human feedback.
\newblock \emph{arXiv preprint arXiv:2305.14975}.

\bibitem[{Venkit et~al.(2024)Venkit, Chakravorti, Gupta, Biggs, Srinath,
  Goswami, Rajtmajer, and Wilson}]{venkit2024confidently}
Pranav~Narayanan Venkit, Tatiana Chakravorti, Vipul Gupta, Heidi Biggs, Mukund
  Srinath, Koustava Goswami, Sarah Rajtmajer, and Shomir Wilson. 2024.
\newblock " confidently nonsensical?'': A critical survey on the perspectives
  and challenges of'hallucinations' in nlp.
\newblock \emph{arXiv preprint arXiv:2404.07461}.

\bibitem[{Vrande\v{c}i\'{c} and Kr\"{o}tzsch(2014)}]{Wikidata}
Denny Vrande\v{c}i\'{c} and Markus Kr\"{o}tzsch. 2014.
\newblock \href {https://doi.org/10.1145/2629489} {Wikidata: a free
  collaborative knowledgebase}.
\newblock \emph{Commun. ACM}, 57(10):78–85.

\bibitem[{Wang et~al.(2023)Wang, Cheng, Guo, Yue, Ding, Xu, Wang, Hu, Zhang,
  and Zhang}]{EM_1}
Cunxiang Wang, Sirui Cheng, Qipeng Guo, Yuanhao Yue, Bowen Ding, Zhikun Xu,
  Yidong Wang, Xiangkun Hu, Zheng Zhang, and Yue Zhang. 2023.
\newblock \href
  {http://papers.nips.cc/paper\_files/paper/2023/hash/f323d594aa5d2c68154433a131c07959-Abstract-Datasets\_and\_Benchmarks.html}
  {Evaluating open-qa evaluation}.
\newblock In \emph{Advances in Neural Information Processing Systems 36: Annual
  Conference on Neural Information Processing Systems 2023, NeurIPS 2023, New
  Orleans, LA, USA, December 10 - 16, 2023}.

\bibitem[{Wei et~al.(2022)Wei, Bosma, Zhao, Guu, Yu, Lester, Du, Dai, and
  Le}]{Instruction-Tuning}
Jason Wei, Maarten Bosma, Vincent~Y. Zhao, Kelvin Guu, Adams~Wei Yu, Brian
  Lester, Nan Du, Andrew~M. Dai, and Quoc~V. Le. 2022.
\newblock \href {https://openreview.net/forum?id=gEZrGCozdqR} {Finetuned
  language models are zero-shot learners}.
\newblock In \emph{The Tenth International Conference on Learning
  Representations, {ICLR} 2022, Virtual Event, April 25-29, 2022}.
  OpenReview.net.

\bibitem[{Yang et~al.(2023)Yang, Chern, Qiu, Neubig, and
  Liu}]{yang2023alignment}
Yuqing Yang, Ethan Chern, Xipeng Qiu, Graham Neubig, and Pengfei Liu. 2023.
\newblock Alignment for honesty.
\newblock \emph{arXiv preprint arXiv:2312.07000}.

\bibitem[{Yin et~al.(2023)Yin, Huang, and Wan}]{ALCUNA}
Xunjian Yin, Baizhou Huang, and Xiaojun Wan. 2023.
\newblock \href {https://doi.org/10.18653/v1/2023.emnlp-main.87} {{ALCUNA}:
  Large language models meet new knowledge}.
\newblock In \emph{Proceedings of the 2023 Conference on Empirical Methods in
  Natural Language Processing}, pages 1397--1414, Singapore. Association for
  Computational Linguistics.

\bibitem[{Yona et~al.(2024)Yona, Aharoni, and Geva}]{GRANOLA}
Gal Yona, Roee Aharoni, and Mor Geva. 2024.
\newblock Narrowing the knowledge evaluation gap: Open-domain question
  answering with multi-granularity answers.
\newblock \emph{arXiv preprint arXiv:2401.04695}.

\bibitem[{Zhang et~al.(2023)Zhang, Diao, Lin, Fung, Lian, Wang, Chen, Ji, and
  Zhang}]{zhang2023r}
Hanning Zhang, Shizhe Diao, Yong Lin, Yi~R Fung, Qing Lian, Xingyao Wang,
  Yangyi Chen, Heng Ji, and Tong Zhang. 2023.
\newblock R-tuning: Teaching large language models to refuse unknown questions.
\newblock \emph{arXiv preprint arXiv:2311.09677}.

\bibitem[{Zhong et~al.(2023)Zhong, Wu, Manning, Potts, and
  Chen}]{DBLP:conf/emnlp/ZhongWMPC23}
Zexuan Zhong, Zhengxuan Wu, Christopher~D. Manning, Christopher Potts, and
  Danqi Chen. 2023.
\newblock \href {https://doi.org/10.18653/V1/2023.EMNLP-MAIN.971} {Mquake:
  Assessing knowledge editing in language models via multi-hop questions}.
\newblock In \emph{Proceedings of the 2023 Conference on Empirical Methods in
  Natural Language Processing, {EMNLP} 2023, Singapore, December 6-10, 2023},
  pages 15686--15702. Association for Computational Linguistics.

\bibitem[{Zhou et~al.(2023)Zhou, Liu, Xu, Iyer, Sun, Mao, Ma, Efrat, Yu, Yu,
  Zhang, Ghosh, Lewis, Zettlemoyer, and Levy}]{LIMA}
Chunting Zhou, Pengfei Liu, Puxin Xu, Srinivasan Iyer, Jiao Sun, Yuning Mao,
  Xuezhe Ma, Avia Efrat, Ping Yu, Lili Yu, Susan Zhang, Gargi Ghosh, Mike
  Lewis, Luke Zettlemoyer, and Omer Levy. 2023.
\newblock \href
  {http://papers.nips.cc/paper\_files/paper/2023/hash/ac662d74829e4407ce1d126477f4a03a-Abstract-Conference.html}
  {{LIMA:} less is more for alignment}.
\newblock In \emph{Advances in Neural Information Processing Systems 36: Annual
  Conference on Neural Information Processing Systems 2023, NeurIPS 2023, New
  Orleans, LA, USA, December 10 - 16, 2023}.

\bibitem[{Zhu et~al.(2020)Zhu, Rawat, Zaheer, Bhojanapalli, Li, Yu, and
  Kumar}]{zhu2020modifying}
Chen Zhu, Ankit~Singh Rawat, Manzil Zaheer, Srinadh Bhojanapalli, Daliang Li,
  Felix Yu, and Sanjiv Kumar. 2020.
\newblock Modifying memories in transformer models.
\newblock \emph{arXiv preprint arXiv:2012.00363}.

\end{thebibliography}
    \bibliographystyle{acl_natbib}
    
    \clearpage

\appendix

\section{Data Preprocessing}
\label{sec:data_prep_appendix}

\begin{table*}[t]
\centering
\resizebox{\textwidth}{!}{%
\begin{tabular}{llrrrrrr}

relation & question template & \knowncat & \maybeknowncat & \weaklyknowncat & \unknowncat & Total & Min \\
\toprule
P131 & Where is [E] located? & 553 & 2529 & 1493 & 3071 & 7646 & 553 \\
P136 & What type of music does [E] play? & 236 & 3410 & 1892 & 1978 & 7516 & 236 \\
P17 & Which country is [E] located in? & 4387 & 2628 & 511 & 364 & 7890 & 364 \\
P19 & Where was [E] born? & 369 & 1884 & 1498 & 4170 & 7921 & 369 \\
P26 & Who is [E] married to? & 1609 & 1503 & 1087 & 3257 & 7456 & 1087 \\
P264 & What music label is [E] represented by? & 206 & 1444 & 1854 & 3820 & 7324 & 206 \\
P36 & What is the capital of [E]? & 4160 & 1634 & 449 & 572 & 6815 & 449 \\
P40 & Who is [E]'s child? & 692 & 1467 & 1271 & 2680 & 6110 & 692 \\
P495 & Which country was [E] created in? & 5459 & 1101 & 408 & 706 & 7674 & 408 \\
P69 & Where was [E] educated? & 233 & 1126 & 1712 & 3650 & 6721 & 233 \\
P740 & Where was [E] founded? & 1323 & 1618 & 1428 & 2902 & 7271 & 1323 \\
P800 & What is [E] famous for? & 301 & 330 & 222 & 503 & 1356 & 222 \\

\midrule
TOTAL & - & 19528 & 20674 & 13825 & 27673 & 81700 & 6142 \\

\bottomrule
\end{tabular}%
}
\caption{Statistics of the \eq train split annotated with \method categories. We annotate the entire train split but always fine-tune on exactly 6142 examples (see the Min column). Refer to \S \ref{sec:controlled_experiment_appendix} for more details.}
\label{tab:train_stats}
\end{table*}

\begin{table*}[t]
\centering
\resizebox{\textwidth}{!}{%
\begin{tabular}{llrrrrr}

relation & question template & \knowncat & \maybeknowncat & \weaklyknowncat & \unknowncat & Total \\
\toprule
P131 & Where is [E] located? & 57 & 362 & 158 & 388 & 965 \\
P136 & What type of music does [E] play? & 6 & 432 & 248 & 281 & 967 \\
P17 & Which country is [E] located in? & 448 & 432 & 65 & 51 & 996 \\
P19 & Where was [E] born? & 107 & 148 & 243 & 501 & 999 \\
P26 & Who is [E] married to? & 177 & 238 & 158 & 378 & 951 \\
P264 & What music label is [E] represented by? & 47 & 157 & 268 & 486 & 958 \\
P36 & What is the capital of [E]? & 580 & 152 & 62 & 86 & 880 \\
P40 & Who is [E]'s child? & 99 & 191 & 167 & 344 & 801 \\
P495 & Which country was [E] created in? & 699 & 147 & 51 & 96 & 993 \\
P69 & Where was [E] educated? & 27 & 145 & 227 & 441 & 840 \\
P740 & Where was [E] founded? & 182 & 245 & 181 & 334 & 942 \\
P800 & What is [E] famous for? & 35 & 50 & 28 & 76 & 189 \\

\midrule
TOTAL & - & 2464 & 2699 & 1856 & 3462 & 10481 \\

\bottomrule
\end{tabular}%
}
\caption{In-distribution test set statistics.}
\label{tab:in_domain_test_stats}
\end{table*}

\begin{table*}[t]
\centering
\resizebox{\textwidth}{!}{%
\begin{tabular}{llrrrrr}

relation & question template & \knowncat & \maybeknowncat & \weaklyknowncat & \unknowncat & Total \\
\toprule
P127 & Who owns [E]? & 125 & 383 & 168 & 314 & 990 \\
P50 & Who is the author of [E]? & 287 & 193 & 115 & 372 & 967 \\
P407 & Which language was [E] written in? & 366 & 153 & 59 & 45 & 623 \\
P176 & Which company is [E] produced by? & 289 & 277 & 181 & 225 & 972 \\
P170 & Who was [E] created by? & 142 & 284 & 120 & 304 & 850 \\
P175 & Who performed [E]? & 94 & 120 & 103 & 663 & 980 \\
P112 & Who founded [E]? & 134 & 116 & 76 & 140 & 466 \\
\midrule
TOTAL & - & 1437 & 1526 & 822 & 2063 & 5848 \\

\bottomrule
\end{tabular}%
}
\caption{Out-of-distribution test set statistics.}
\label{tab:out_of_domain_test_stats}
\end{table*}

This section expands \S \ref{sec:exp_setting} with additional details about our data preprocessing steps.
The \eq dataset \cite{Entity_Questions} consists of train, development and test splits and spans 24 relations. Our train, development and test sets are curated based on the original splits from \eq.
However, we use only 12 relations, since we wanted to reserve some relations for out-of-distribution test set.
To avoid cherry-picking, the 12 relations used in our train, development and test sets are randomly sampled. The resulting relations are presented in Tables \ref{tab:train_stats} and \ref{tab:in_domain_test_stats}. 

We reserved the remaining 12 relations for out-of-distribution test set. However, we found that in those 12 reserved relations, 5 were too similar to some of the relations that we train on (Table \ref{tab:train_stats}), thus we suspected that this could lead to a test set that is not truly out-of-distribution. To address that, we filtered out those relations and were left with 7 relations for our-of-distribution. These 7 out-of-distribution relations are presented in \Cref{tab:out_of_domain_test_stats}. The relations that were filtered-out are as follows:

\begin{itemize}
    \item P276 was filtered out since it directly overlaps with P131 since for both relations the question in \eq is of the form \textit{``Where is [E] located?''}. 
    P276 stands for ``location'' (\url{https://www.wikidata.org/wiki/Property:P276}) and P131 stands for ``located in the administrative territorial entity'' (\url{https://www.wikidata.org/wiki/Property:P131}).
    
    \item P20, for which the question template is \emph{``Where did [E] die?''}, was filtered out since it may require knowledge that relates to
    P19, for which the question template is \emph{``Where was [E] born?''}.
    P20 stands for ``place of death'' (\url{https://www.wikidata.org/wiki/Property:P20}) and P19 stands for ``place of birth'' (\url{https://www.wikidata.org/wiki/Property:P19}).
    
    \item P106, for which the question template is \emph{``What kind of work does [E] do?''}, was filtered out since it may require knowledge that relates to
    P800, for which the question template is \emph{``What is [E] famous for?''}.
    P106 stands for ``occupation'' (\url{https://www.wikidata.org/wiki/Property:P106}) and P800 stands for ``notable work'' (\url{https://www.wikidata.org/wiki/Property:P800}).
    
    \item P413, for which the question template is \emph{``What position does [E] play?''}, was filtered out since it may require knowledge that relates to
    P800, for which the question template is \emph{``What is [E] famous for?''}.
    P413 stands for ``position played on team / speciality'' (\url{https://www.wikidata.org/wiki/Property:P413}) and P800 stands for ``notable work'' (\url{https://www.wikidata.org/wiki/Property:P800}).
    
    \item P159, for which the question template is \emph{``Where is the headquarters of [E]?''}, was filtered out since it may require knowledge that relates to
    P36, for which the question template is \emph{``What is the capital of [E]?''}.
    P159 stands for ``headquarters location'' (\url{https://www.wikidata.org/wiki/Property:P159}) and P36 stands for ``capital'' (\url{https://www.wikidata.org/wiki/Property:P36}).
\end{itemize}

Lastly, we perform two additional filtering steps: (1) To simplify the process of categorizing the examples w.r.t. $M$'s knowledge (\S \ref{sec:categorizing}), we filter-out examples with more than 1 correct answer.\footnote{$4.2\%$ and $3.9\%$ of the \eq train and test set respectively.} (2) We make sure that no subjects or objects overlap between the train and test sets,\footnote{
For example, the subject \textit{``\textbf{Bruce Smith}''} appears with 2 different relations ($P106$ and $P413$) yielding 2 examples: (\textit{``What kind of work does \textbf{Bruce Smith} do?''}, \textit{``poet''}) and (\textit{``Where was \textbf{Bruce Smith} born?''}, \textit{``Faribault''}).} by filtering-out overlapping examples from the train set.\footnote{$2.1\%$ of the \eq train set.} 

\section{Hallucinations in the Context of our Study}
\label{sec:hallucinations_vs_perfromance_appendix}

In general, the term ``hallucinations'' is not well-defined in NLP \cite{venkit2024confidently}. For clarity, in \S \ref{app:hallucinations} we define the type of hallucinations we target in this work. In addition, while our main takeaway is that acquiring new knowledge can lead to hallucinations, our experiments focus on measuring accuracy drops on the test set. Therefore, in \S \ref{app:hallucinations_and_performacne} we clarify why worse performance on the test set in our study is attributed to higher rate of hallucinations, as defined in \S \ref{app:hallucinations}.

\subsection{Hallucinations w.r.t. the Pre-existing Knowledge.}
\label{app:hallucinations}
\citet{hallucinations_survey} categorized hallucinations into two main categories: (1) \emph{factuality hallucination} and (2) \emph{faithfulness hallucination}.
The first case refers to factual inconsistency between the generated content and  verifiable real-world facts. Common examples include wrong answers in closed-book QA setting \cite{chern2023factool} or factual mistakes in long-form generations of knowledge intensive passages such as biographies \cite{manakul2023selfcheckgpt,min-etal-2023-factscore}. On the other hand, \emph{faithfulness hallucination} refers to cases where the generated content is factually inconsistent with the context provided by the input. A common example is when the model summarizes a document and the resulting summary is factually inconsistent with the input document \cite{honovich-etal-2022-true-evaluating,laban-etal-2022-summac,honovich-etal-2021-q2,gekhman-etal-2023-trueteacher,scialom-etal-2021-questeval,kryscinski-etal-2020-evaluating}.

In this work we focus on a subset of \emph{factuality hallucinations}. Our goal is to study how introducing new factual knowledge through fine-tuning affects the utilization of the model's pre-existing knowledge. To reflect this, we define \emph{hallucinations w.r.t the model’s pre-existing knowledge} as $(q,a)$ pairs that were known to the model after pre-training (as defined by \method), while the fine-tuned model fails to answer $q$ correctly post fine-tuning.\footnote{This can happen due to 2 main reasons: (1) The model still encodes some knowledge regarding the answer to $q$ but hallucinates. (2) The model completely \emph{forgets} the answer to $q$ during fine-tuning. In this work, we treat these two cases collectively as \emph{hallucinations w.r.t the pre-existing knowledge}. Methodologically studying forgetting during fine-tuning and whether unknown facts are still captured in the model’s weights can be an interesting direction for future research.}

\subsection{Test performance as Proxy for Hallucinations.}
\label{app:hallucinations_and_performacne}
We now detail the relation between the test performance in our setting and hallucinations.
In our study, poorer performance of a fine-tuned model $M_{D1}$, compared to another fine-tuned model $M_{D2}$ on the test set, can be attributed to a higher rate of hallucinations in $M_{D1}$, relative to its pre-existing knowledge, due to the following explanation.

The test set can be conceptually divided into two types of questions. First, there are questions with answers that are unknown to $M$. Those questions will remain unknown post fine-tuning, as we make sure that the training set is disjoint from the test set (\S \ref{sec:data_prep_appendix}). This means that both $M_{D1}$ and $M_{D2}$ will fail to answer these questions. Thus, the test performance difference between $M_{D1}$ and $M_{D2}$ is mostly attributed to the second type of questions: ones that are known to $M$, i.e. $M$ can answer them correctly since it posses the relevant knowledge. 
Thus, $M_{D1}$ and $M_{D2}$ must rely on their pre-existing knowledge to answer such questions, and a lower performance on such question can be only categorized as an hallucination w.r.t. pre-existing knowledge.

\section{$\boldsymbol{P_{\boldsymbol{\mathtt{Correct}}}}$ Approximation}
\label{sec:score_measure_appendix}

This section expands \S \ref{sec:categorizing} with additional details about our $\score$ approximation.
In our study we approximate $\score(q, a; M, T)$ based on the fraction of correct answers to $q$ sampled from $M$.
We begin with randomly sampling $N_{\text{ex}}$ distinct $k$-shot exemplars for each relation in our dataset (\S \ref{sec:data_prep_appendix}).
Then, to approximate $\score(q, a; M, T)$, we use $M$ to generate answers to $q$ using each of the $N_{\text{ex}}$ exemplars from the relation corresponding to $q$. 
We first use temperature sampling with $T=0.5$ to sample $N_{\text{sample}}$ answers for each of the $N_{\text{ex}}$ exemplars.
$\score(q, a; M, T>0)$ is then approximated by the fraction of correct answers from the total of $N_{\text{ex}}\cdot N_{\text{sample}}$ predictions. We also generate the greedy decoding prediction ($T=0$) for each of the $N_{\text{ex}}$ exemplars. $\score(q, a; M, T=0)$ is then approximated by the fraction of correct answers from the total of $N_{\text{ex}}$ predictions.\footnote{Since we can only have one greedy prediction for every k-shot exemplars.} 

We use $k=4$ in our study, simply since we found it enough for $M$ to output answers in the correct format. 
We use $N_{\text{ex}}=10$ and $N_{\text{sample}}=16$. The $N_{\text{sample}}=16$ samples using $T=0.5$ are sampled from Top 40.

The $k$ exemplars are sampled from the development split. We sample $N_{\text{ex}}$ different samples since we found that even when the few-shot exemplars are sampled per-relation, their exact choice still affects the prediction. In \S \ref{sec:taxonomy} and \Cref{fig:ptrue_eval} we show evidence that this also improves the quality of our categories.

Below is an example of our 4-shot prompt format, from real example from \eq with the relation $P106$ representing occupation.\footnote{\url{https://www.wikidata.org/wiki/Property:P106}} The question in this case is \emph{``What kind of work does Ron Konopka do?''} and the ground truth asnwer is \emph{``geneticist''}. 

\begin{mdframed}[backgroundcolor=blue!5, skipabove=0.5\baselineskip]
\small

\noindent Q: What kind of work does Nicolas Roeg do?

\noindent A: film director

\noindent Q: What kind of work does Crystal Geoffré do?

\noindent A: actor

\noindent Q: What kind of work does Maurice Blondel do?

\noindent A: philosopher

\noindent Q: What kind of work does Javier de Burgos do?

\noindent A: politician

\noindent Q: What kind of work does Ron Konopka do?

\noindent A:


\end{mdframed}
\vspace{0.5\baselineskip}

\begin{table}[t]
\centering
\resizebox{\columnwidth}{!}{%
\begin{tabular}{llll}

\multicolumn{1}{l}{\multirow{1}{*}{Wrong Answer}}  &
  \multicolumn{1}{c}{\text{Paraphrase}} &
  \multicolumn{1}{c}{\text{Higher Granularity}} &
  \multicolumn{1}{c}{\text{Lower Granularity}}\\
  \toprule

$90\%$    & $6\%$
 & $2\%$  & $2\%$ \\

\bottomrule
\end{tabular}%
}
\caption{
Error Analysis of 100 Predictions of the Pre-trained Model, for Which Exact Match is False.
}
\label{tab:error_analysis}
\end{table}

To decide whether a sampled answer is correct, we use the Exact Match (EM) metric to compare it with the ground truth answer. The main advantage in this choice is that when EM is True, we know that the answer is correct for $100\%$. The main potential risk associated with this choice is that we may wrongly classify answers as incorrect due to paraphrases or answers with different granularity levels \cite{EM_1, EM_2, GRANOLA}). To address this, we perform an \textbf{error analysis} on 100 predictions for which EM was False. We randomly sample  50 greedy predictions ($T=0$) and 50 samples with $T=0.5$. The results are in \Cref{tab:error_analysis}. This analysis suggest  that in $90\%$ of the cases where EM is False, the predicted answer is indeed incorrect. Which is a reasonable performance for our purpose, especially considering that when EM is True the answer is $100\%$ correct.

\section{Data Annotation}
\label{sec:data_annotation_appendix}

we first calculate $\score(q,a;M,T=0)$ and $\score(q,a;M,T>0)$ for each $(q,a)$ pair in our preprocessed dataset (\S \ref{sec:exp_setting} and \S \ref{sec:data_prep_appendix}), using our $\score(\cdot)$ approximation (\S \ref{sec:categorizing} and \S \ref{sec:score_measure_appendix}). We then use these values to categorize each $(q,a)$ pair into one of our four categories (\S \ref{sec:categorizing} and \Cref{tab:categories_def}). We provide the full statistics of the categories on the train and test set, as well as the out-of-distribution test set in Tables \ref{tab:train_stats}, \ref{tab:in_domain_test_stats} and \ref{tab:out_of_domain_test_stats}.

\section{Fine-tuning Details}
\label{sec:controlled_experiment_appendix}

\paragraph{Fine-tuning Data.}In \S \ref{sec:results} we examine the effect of new knowledge in the fine-tuning dataset $D$ on the performance of $M_D$, by varying the proportion of \unknowncat examples in $D$.
When we create variants of $D$ with exactly $X\%$ of \unknowncat and $(100-X)\%$ \knownSolo examples, 
we make sure that the relation distribution remains consistent. To achieve that we sample $X\%$ of \unknowncat \emph{from each relation}.

In \S \ref{sec:single_cat} we create single-category variants of $D$. Since we want to work with a fixed $|D|$ across all variants, we want to make sure that we have $|D|$ examples from each category. To ensure this, we measure the size of the smallest category in each relation (see the ``Min'' column in \Cref{tab:train_stats}) and define $|D|$ as their sum. 
In other words, for each relation we calculate the size of the smallest category and sum these values. This leads to $|D|=6142$, as illustrated by the last column in \Cref{tab:train_stats}.
More formally, for each relation $\text{r}$ in the training split, and for each category CAT from our 4 \method categories, we define $\text{CAT}_\text{r}$ to be the examples from category $\text{CAT}$ and relation $\text{r}$. Consequently $\text{size}(\text{CAT}_\text{r})$ is the number of the examples in $\text{CAT}_\text{r}$. For example 
\mbox{$\text{size}($\knowncat$_{\text{P131}}) = 553$}
(see \Cref{tab:train_stats}). We then define:
\[
|D| = \sum_{r \in R_{\text{Train}}} \min \left\{ \text{size}(CAT_{r}) | \ 
\left.
\begin{array}{l}
\text{\scriptsize CAT} \in \{ \\
\text{\scriptsize \knowncat}, \\
\text{\scriptsize \maybeknowncat}, \\
\text{\scriptsize \weaklyknowncat}, \\
\text{\scriptsize \unknowncat}\} \\
\end{array}
\right\} \right.
\]
where $\text{R}_\text{Train}$ are the 12 relations from the training set.

Below is an example of our data format in the train, development and test sets, from real example from \eq with the relation $P106$ representing occupation.\footnote{\url{https://www.wikidata.org/wiki/Property:P106}} The question in this case is \emph{``What kind of work does Ron Konopka do?''} and the ground truth asnwer is \emph{``geneticist''}. 
\begin{mdframed}[backgroundcolor=blue!5, skipabove=0.5\baselineskip]
\small

\noindent Answer the following question.

\noindent What kind of work does Ron Konopka do?


\end{mdframed}
\vspace{0.5\baselineskip}

\paragraph{Fine-tuning Regime.}
In this work, we focus on full fine-tuning, where all model parameters are updated. An interesting direction for future research is to investigate similar questions within parameter-efficient fine-tuning regimes \cite{han2024parameter}, such as LoRA \cite{DBLP:conf/iclr/HuSWALWWC22}. For instance, \citet{biderman2024lora} demonstrated that, compared to full fine-tuning, LoRA better preserves the base model’s performance on tasks outside the target domain, though at the cost of diminished performance within the target domain. It could be interesting to check if this also holds to hallucinations w.r.t. the models pre-existing knowledge as we define it in this work (\S \ref{sec:hallucinations_vs_perfromance_appendix}). Another interesting avenue for future research is to explore how new knowledge is acquired during continual pre-training \cite{jiang-etal-2024-instruction,parmar2024reuse,ibrahim2024simple} as one of the key objectives in continual pre-training is to inject new (up to date) knowledge to the model.

\begin{figure}[t]
 \centering
\vspace{-1.5mm} 
 \includegraphics[width=\columnwidth]{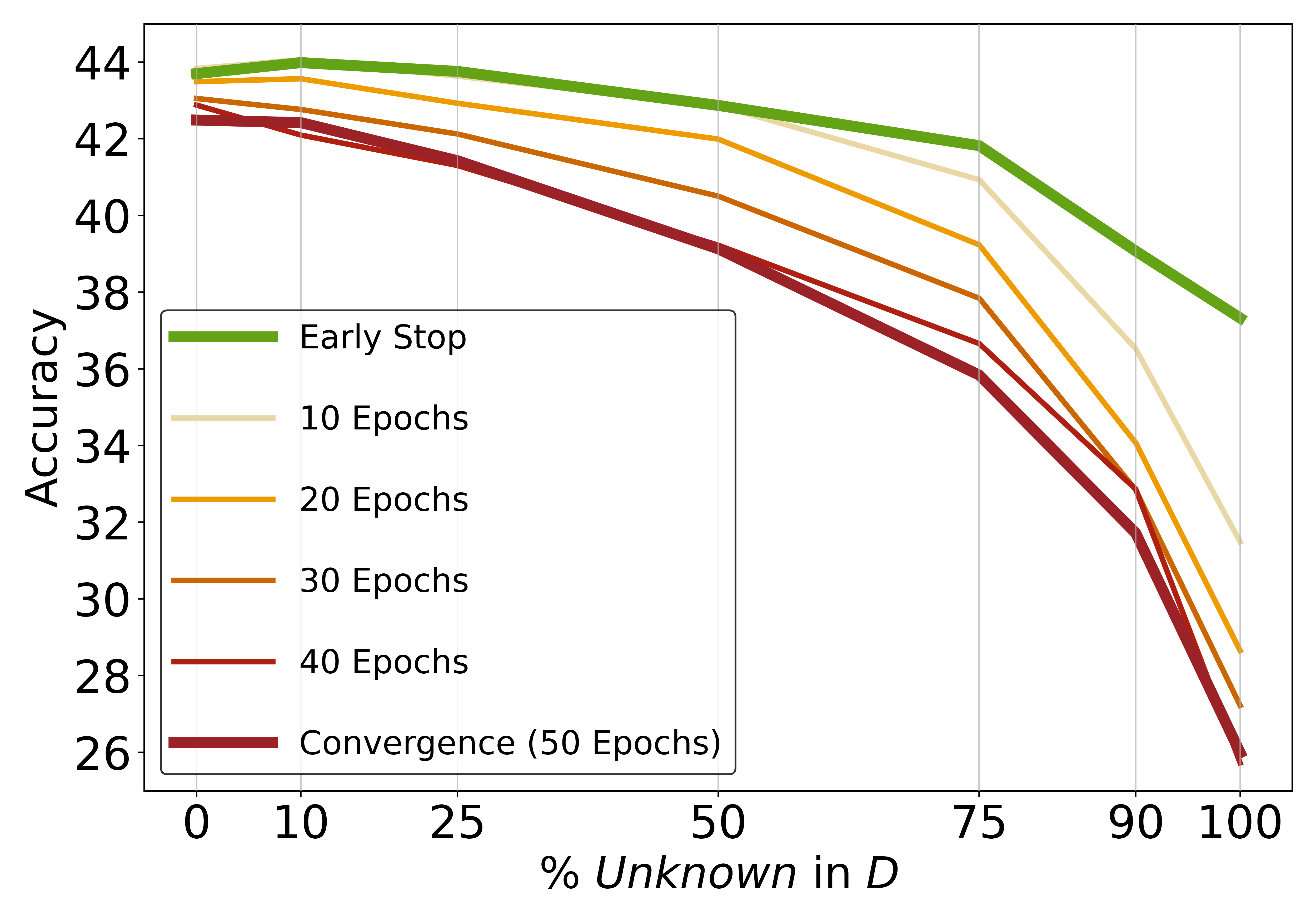}   
    \caption{
    Performance on the test set with a slower learning rate of 1e-4. This plot is equivalent to \Cref{fig:fixed_steps_plot}, and the results are similar, except that the experiment were run with a learning rate of 1e-4 instead of 1e-5.
    }
    \label{fig:fixed_steps_plot_lr04}    
\end{figure}

\paragraph{Fine-tuning hypeparameters.} We fine-tune every model for 50 epochs for all our model variants to completely fit the training set, so we can examine all stages of fine-tuning.
We evaluate the models every epoch on the development set. 
The \maxdev stopping criteria is defined to be the epoch with the maximum accuracy on the development set.
We use learning rate of 1e-5, a batch
size of 128, and a dropout rate of 0.05.
Our experimental design intentionally utilized a fixed learning rate, which is the standard for supervised fine-tuning, as opposed to the dynamic learning rate strategies typically employed in continual pre-training. 
We have experimented with both slower and faster fixed learning rates (1e-4 and 1e-6) to ensure the robustness of our conclusions. These experiments consistently supported our findings.
For instance, in \Cref{fig:fixed_steps_plot_lr04} we present the performance as a function of the $\%$ of the \unknowncat examples in $D$ (i.e. similar plot to \Cref{fig:fixed_steps_plot}) when using a learning rate of 1e-4 instead of 1e-5.

\section{The Case for Avoiding Fake Facts}
\label{sec:fake_facts_appendix}

One limitation of using the \unknowncat examples in our study is that \method only approximates the LLM's knowledge. This means that some examples can be incorrectly classified as unknown to $M$. As we discuss in \S \ref{sec:taxonomy}, our results indicate that this happens in at most $3\%$ of the cases, meaning that the vast majority of the examples classified as \unknowncat are actually unknown to $M$.

Alternative approach could be to simply use fake facts as unknown fine-tuning examples. We considered this in early stages of the project and were concerned that this would introduce confounding factors into our study, as fake facts may behave differently than real ones; In our setup where the knowledge is represented with \emph{(subject, relation, object)} triplets, there are 2 main ways to generate fake facts: (1) creating triplets where both the \emph{subject} and the \emph{object} are fake \cite{ALCUNA}. (2) Creating triplets where the \emph{subject} is real and the \emph{object} is fake \cite{zhu2020modifying,DBLP:conf/nips/MengBAB22,DBLP:conf/iclr/MengSABB23,DBLP:conf/emnlp/ZhongWMPC23}. Focusing exclusively on (1) will capture only a small subset of the cases of new factual knowledge, as in the majority of the cases the subject will be familiar to the pre-trained model. E.g., the model may know that there is a person named \textit{``Barack Obama''} but not know where he was born. If we consider (2), using real subjects with fake objects may compromise our study as in many cases this will introduce \textbf{knowledge updates} and not new knowledge. To illustrate this, let’s consider the triplet \textit{(``Barack Obama'', ``place of birth'', \textbf{``Honolulu''})}, and let’s assume that we generate the fake triplet  \textit{(``Barack Obama'', ``place of birth'', \textbf{``London''})}. Now, since the original (correct) triplet may be known to the model we essentially do not simulate introducing new factual knowledge but \textbf{updating existing knowledge}. Considering the above, we decided that using real world facts will make our findings more reliable. We then invested a considerable effort to ensure that the examples that are classified as unknown are truly unknown to the model (as discussed above).

\section{Train Accuracy on Different \knownSolo Categories}
\label{sec:ltraining_accuracy_appendix}

\begin{figure}[t]
 \centering
  \vspace{-0.8cm}
\includegraphics[width=\columnwidth]{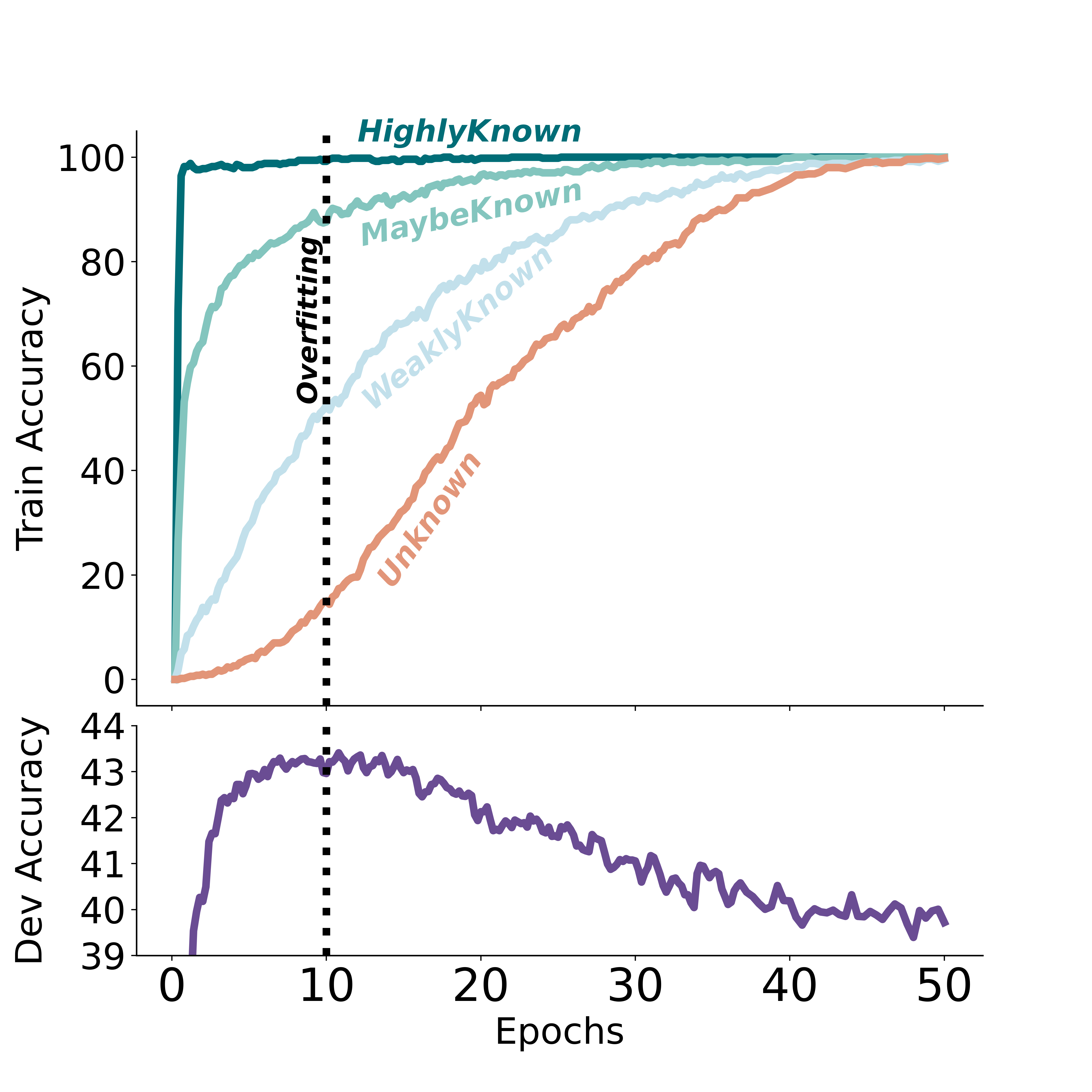}   
    \caption{
    Training accuracy as a function of fine-tuning duration, evaluated on the variant with $50\%$ \unknowncat fine-tuning examples. 
    For reference, we also include the accuracy on the development set, accompanied by a zoom-in plot within a narrower range, to provide a more visible and clear view.
    }
    \label{fig:training_accuracy_plot_w_dev}    
\end{figure}

\begin{figure*}[t]
 \centering
 \begin{subfigure}{.5\textwidth}
  \centering
\includegraphics[width=\linewidth]{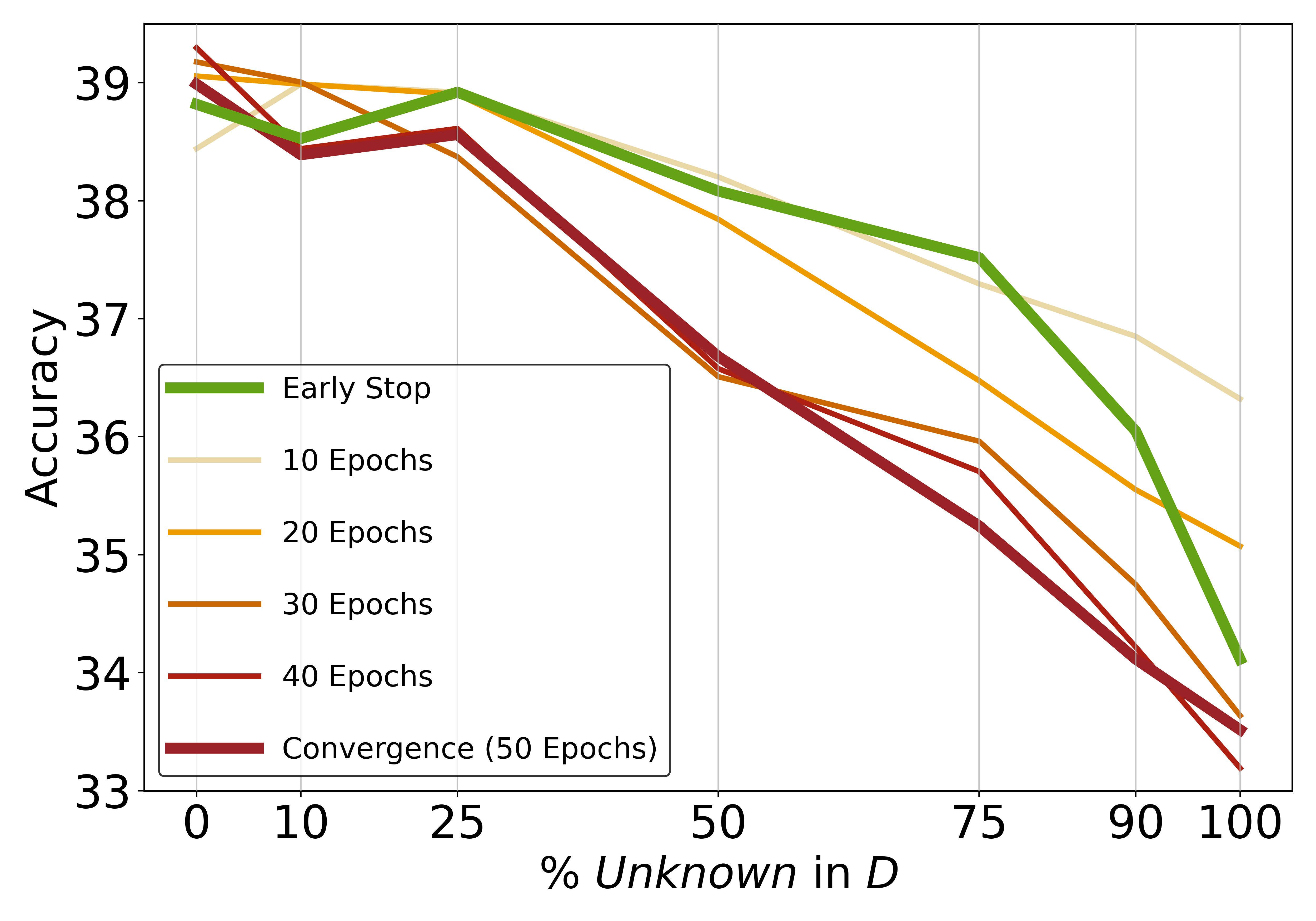}
  \caption{
  }
  \label{fig:fixed_steps_plot_ooo}
 \end{subfigure}%
 \begin{subfigure}{.5\textwidth}
  \centering
\includegraphics[width=\linewidth]{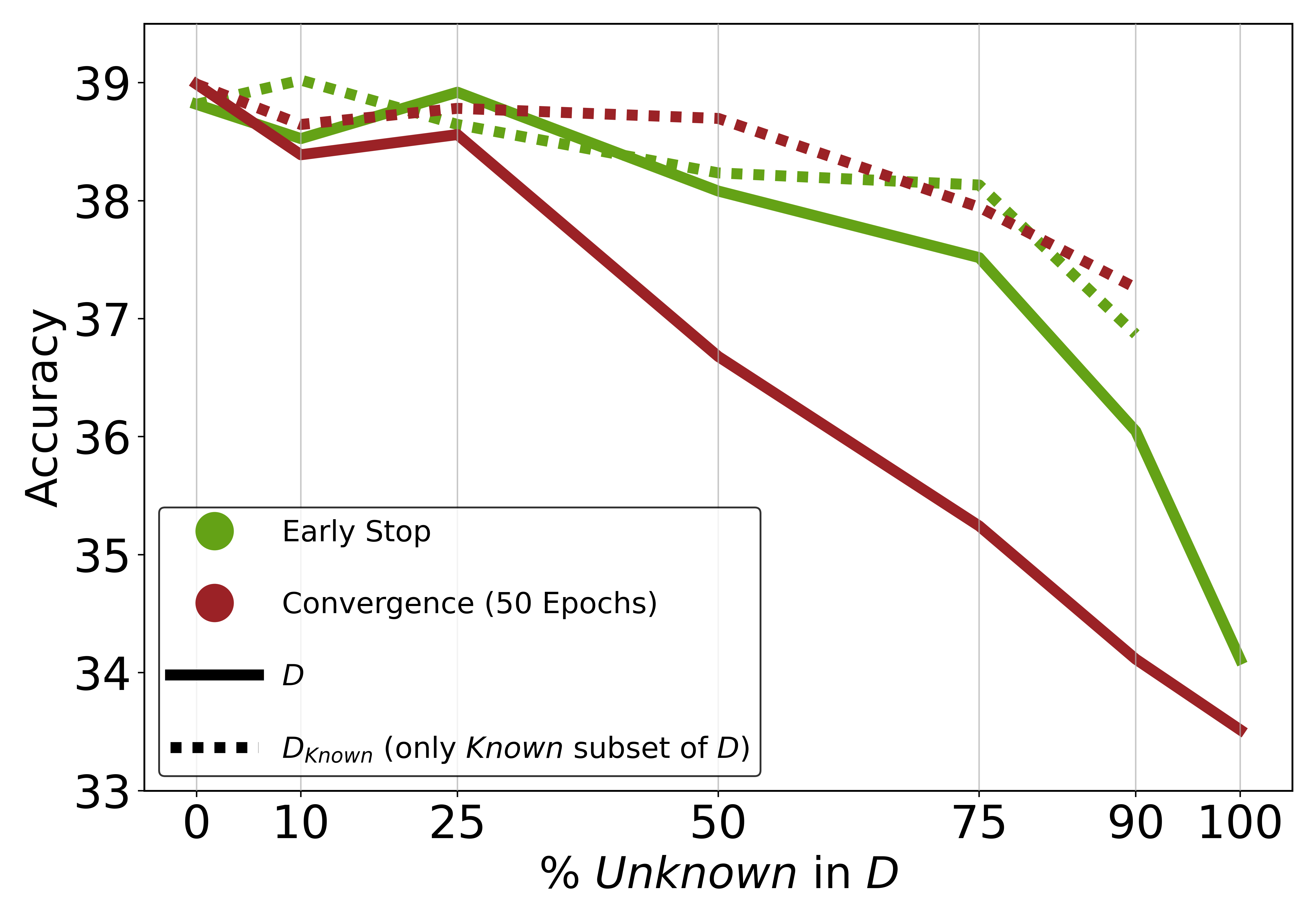}
  \caption{
  }
  \label{fig:ablated_plot_ooo}
 \end{subfigure}
 \caption{
  Performance on the \emph{out-of-distribution (OOD)} test set as a function of the $\%$ of \unknowncat examples in the fine-tuning dataset $D$. This plot is the OOD version of \Cref{fig:combined}. Everything is similar to \Cref{fig:combined}, except that y-axis is the accuracy on the OOD test set. We note that \emph{\textbf{the development set did not change (not OOD)}}, thus it does not necessarily reflects the optimal stopping point for OOD.
 }
 \label{fig:combined_ood}
\end{figure*}

In \S \ref{sec:ignore} we analyze the fine-tuning dynamic and present the training accuracy as function of the fine-tuning duration in \Cref{fig:main_plot}. For simplicity we treated the \knownSolo categories collectively. For reference we also include the plot with the full per-category breakdown in
\Cref{fig:training_accuracy_plot_w_dev}.

\section{Linear Model}
\label{sec:linear_regression_appendix}
In \S \ref{sec:equation} and \S \ref{sec:ood} we use a linear model (\Cref{eq:accuracy}) that predicts that test accuracy and the out-of-distribution test accuracy.
We estimate the parameters of this linear model based on results from all our variants of $D$ used in \S \ref{sec:results}.
For all these variants, we measure the test accuracy and the number of \knownSolo and \unknowncat fine-tuning examples that $M$ fits during different fine-tuning stages. This way we collect a dataset with examples of the form $(Accuracy, N_\text{Kn}, N_\text{Unk})$, which we use to fit a linear regression model.

\section{Out-of-distribution (OOD) Evaluation}
\label{sec:ood_appendix}

In \S \ref{sec:ood} we discuss \emph{out-of-distribution (OOD)} results. In these experiments we simply used our OOD test set consisting of 7 relations unseen during fine-tuning (see \S \ref{sec:data_prep_appendix}). When we perform the analysis discussed in \S \ref{sec:higher_unk_affects_performance} and \S \ref{sec:harmful_or_neutra}, we additionally evaluated the models on the OOD test set. For completeness, we add here \Cref{fig:combined_ood}, which is the out-of-distribution version of \Cref{fig:combined}. 
\Cref{fig:fixed_steps_plot_ooo} presents the OOD test performance as a function of $\%$ of \unknowncat examples in $D$ for different fine-tuning duration. The corresponding \emph{in-distribution} results (\Cref{fig:fixed_steps_plot}) were discussed in \S \ref{sec:higher_unk_affects_performance}.
\Cref{fig:ablated_plot_ooo} presents the OOD test performance for the ablation where we filter-out \unknowncat fine-tuning examples. The corresponding \emph{in-distribution} results (\Cref{fig:ablated_plot}) were discussed in \S \ref{sec:harmful_or_neutra}.
We notice that similar trends, just with a smaller overall magnitude of the performance drop, up to 6 points drop compared to up to 14 for in-distribution. This smaller drop magnitude is also reflected in smaller values of $|\beta_\text{ukn}|$ and $|\beta_\text{kn}|$ (\Cref{tab:linear_model}).

\begin{table*}[t]
\centering
\resizebox{\textwidth}{!}{%
\begin{tabular}{llcllllclcllll}

  &
\multicolumn{6}{c}{\maxdev} &
\phantom{abcd} &
\multicolumn{6}{c}{\conv} \\

\cmidrule{2-7} \cmidrule{9-14} 

  &
  \multicolumn{1}{c}{$\mathtt{Full}$} &
  \phantom{a}&
  \multicolumn{1}{c}{\knowncatshort} &
  \multicolumn{1}{c}{\maybeknowncatshort} &
  \multicolumn{1}{c}{\weaklyknowncatshort} &
  \multicolumn{1}{c}{\unknowncatshort} &&
  \multicolumn{1}{c}{$\mathtt{Full}$} &
  \phantom{a}&
  \multicolumn{1}{c}{\knowncatshort} &
  \multicolumn{1}{c}{\maybeknowncatshort} &
  \multicolumn{1}{c}{\weaklyknowncatshort} &
  \multicolumn{1}{c}{\unknowncatshort} \\
\toprule

\known & 40.5$^{**}$  && \textbf{98.7} & 60.1$^{**}$ & 9.0$^{**}$ & 0.6$^{**}$ && 40.0$^{**}$ &&  \textbf{98.4} & 58.8$^{**}$ & 8.5$^{**}$ & 0.7$^{**}$ \\

\maybeknown & \textbf{43.6}  && \textbf{98.4} & \textbf{69.9} & 12.1$^{**}$ & 1.0$^{**}$ && \textbf{43.2}  && 97.5$^{*}$ & \textbf{68.2} & 12.9$^{**}$ & 1.3$^{**}$ \\

\weaklyknown & 39.2$^{**}$  && 95.0$^{**}$ & 59.2$^{**}$ & 8.6$^{**}$ & 0.4$^{**}$ && 35.4$^{**}$  && 73.5$^{**}$ & 55.8$^{**}$ & \textbf{17.2} & 2.2$^{**}$ \\

\unknownfull & 37.5$^{**}$ &&  95.6$^{**}$ & 52.9$^{**}$ & 6.5$^{**}$ & 0.6$^{**}$ && 25.8$^{**}$  && 55.8$^{**}$ & 36.6$^{**}$ & 12.2$^{**}$ & \textbf{3.2} \\

\random & \textbf{43.5}  && 98.0$^{*}$ & 67.6$^{**}$ & \textbf{14.1} & \textbf{1.8} && 41.8$^{**}$ && 95.5$^{**}$ & 61.7$^{**}$ & 14.8$^{**}$ & 2.5$^{*}$ \\

\bottomrule
\end{tabular}%
}
\caption{
A copy of \Cref{tab:categories_analysis} with detailed notation of the statistic significant test results. In each column, statistically significant differences from the best result are indicated using $^{*}$ and $^{**}$ for \(p < 0.05\) and \(p < 0.01\) respectively.
}
\label{tab:categories_analysis_with_stat_sig}
\end{table*}

\section{Statistic Significance Tests}
\label{sec:stat_sig_appendix}

In \S \ref{sec:single_cat} we present \Cref{tab:categories_analysis}. As mentioned in the caption, we perform statistic significance tests for each column. To this end we compare all the values to the maximal value in this column. 

For each subset of the test set, we randomly shuffle all the examples in it, split them up into 100 approximately equally sized subsets, and compute accuracy for each of them for all the models of interest. 
We then apply paired-sample t-test with $p < 0.05$ and $p < 0.01$.

In \Cref{tab:categories_analysis}, the best result is in bold, as well as all the results with statistically non-significant difference from the best with $p < 0.05$. We additionally include a copy of \Cref{tab:categories_analysis} where all the statistical tests outcomes are annotated, see \Cref{tab:categories_analysis_with_stat_sig}.
We can see that in almost all cases the difference is statistically significant with $p < 0.01$, except two cases where it is only with $p < 0.05$ (\random \unknowncatshort\xspace and \maybeknown \maybeknowncatshort).

Since we also discuss ``horizontal'' comparisons, where we compare \maxdev to \conv, we additionally run significance tests (not annotated in \Cref{tab:categories_analysis}) for $All$, comparing \maxdev to \conv. The difference for \maybeknown was not statistically significant while for all others (including \random) it was significant with $p < 0.01$.

\section{The P(True) Case Study}
\label{sec:p_true_appendix}

In \S \ref{sec:taxonomy} we used the P(True) metric from \citet{p_true} as a case study for comparison. In \Cref{fig:ptrue_eval} we compare our \unknowncat category vs classifying as \unknowncat based on a threshold of P(True). We calculated P(True) for every $(q,a)$ pair in the test set using \citet{p_true}'s prompt:

\begin{mdframed}[backgroundcolor=blue!5, skipabove=0.5\baselineskip]
\small
\noindent Question: \textit{Where is Paris located?}

\noindent Proposed Answer: \textit{France}

\noindent Is the proposed answer:

(A) True

(B) False

\noindent The proposed answer is:

\end{mdframed}
\vspace{0.5\baselineskip}
We then treated $(q,a)$ pairs with P(True) below a threshold as \unknowncat.
We experimented with each possible threshold $T$ in $[0,1]$, according to our test set. For each threshold $T$ we then measured (1) how many examples were classified as \unknowncat out of the test set, (2) what was the accuracy on these examples after fine-tuning. We plot the results in \Cref{fig:ptrue_eval}, where P(True) is represented with the \textcolor{ptruecolor}{\textbf{yellow line}} and our \unknowncat is represented with the \textcolor{ourscolor}{\textbf{blue circle}}. 
As discussed in \S \ref{sec:score_measure_appendix}, it was approximated using 10 defferent samples of 4-shot exemplars ($N_{\text{ex}}=10$). We also check smaller values of $N_{\text{ex}}$ and plot the results with the \textcolor{ourscolor}{\textbf{blue line}}. 
The accuracy after fine-tuning for all the results is measured after fine-tuning with \random (\S \ref{sec:single_cat}).

\end{document}